\documentclass{article}

\usepackage[a4paper,total={6in, 8in}]{geometry}

\usepackage[utf8]{inputenc} 
\usepackage[T1]{fontenc}    
\usepackage{hyperref}       
\usepackage{url}            
\usepackage{booktabs}       
\usepackage{amsfonts}       
\usepackage{nicefrac}       
\usepackage{microtype}      
\usepackage{amsmath}

\usepackage{algorithm}
\usepackage{algpseudocode}
\usepackage{cite}
\usepackage{natbib}

\usepackage{graphicx}
\usepackage{subcaption}

\newtheorem{proposition}{Proposition}

\title{Deep Metric Tensor Regularized Policy Gradient}

\author{Gang Chen\\ School of Engineering and Computer Science\\
  Victoria University of Wellington\\
  New Zealand\\
  \texttt{aaron.chen@ecs.vuw.ac.nz} \\
\and  Victoria Huang\\
  National Institute of Water and Atmospheric Research\\
  New Zealand\\
  \texttt{Victoria.Huang@niwa.co.nz}
 }

\begin{document}

\maketitle

\begin{abstract}
Policy gradient algorithms are an important family of deep reinforcement learning techniques. Many past research endeavors focused on using the first-order policy gradient information to train policy networks. Different from these works, we conduct research in this paper driven by the believe that properly utilizing and controlling Hessian information associated with the policy gradient can noticeably improve the performance of policy gradient algorithms. One key Hessian information that attracted our attention is the Hessian trace, which gives the divergence of the policy gradient vector field in the Euclidean policy parametric space. We set the goal to generalize this Euclidean policy parametric space into a general Riemmanian manifold by introducing a metric tensor field $g_{ab}$ in the parametric space. This is achieved through newly developed mathematical tools, deep learning algorithms, and metric tensor deep neural networks (DNNs). Armed with these technical developments, we propose a new policy gradient algorithm that learns to minimize the absolute divergence in the Riemannian manifold as an important regularization mechanism, allowing the Riemannian manifold to smoothen its policy gradient vector field. The newly developed algorithm is experimentally studied on several benchmark reinforcement learning problems. Our experiments clearly show that the new metric tensor regularized algorithm can significantly outperform its counterpart that does not use our regularization technique. Additional experimental analysis further suggests that the trained metric tensor DNN and the corresponding metric tensor $g_{ab}$ can effectively reduce the absolute divergence towards zero in the Riemannian manifold.
\end{abstract}

\section{Introduction}
\label{sec-int}

Policy gradient methods are an important family of \emph{deep reinforcement learning} (DRL) algorithms. They help a DRL agent find an \emph{optimal policy} that maps any states the agent encounters to optimal actions \cite{schulman2017proximal,lillicrap2015continuous}. Unlike Q-learning and other value-based methods, policy gradient methods directly learn a \emph{deep neural network} (DNN) known as a \emph{policy network} \cite{sutton2000policy,lillicrap2015continuous}. This is achieved by computing the \emph{policy gradient} with respect to the trainable parameters of the policy network, known as \emph{policy parameters}, and updating the parameters in the direction of optimizing an agent's \emph{expected cumulative return}. For this purpose, several key techniques are often used jointly, including Monte Carlo simulation \cite{llorente2021survey}, gradient estimation \cite{sutton2000policy}, and optimization based on \emph{stochastic gradient descent} (SGD) \cite{goodfellow2016deep}.

Existing research showed that the accuracy of policy gradient has a profound impact on the performance of DRL algorithms \cite{fujimoto2018,wang2020striving_sop,lee2021sunrise}. In view of this, substantial efforts have been made previously to reduce the bias and variance of the estimated policy gradient \cite{haarnoja2018, fan2021explaining,ZHANG202040}. Ensemble learning and hybrid on/off-policy algorithms have also been developed to facilitate reliable estimation of policy gradients for improved exploration and sample efficiency \cite{lee2021sunrise,januszewski2021,chen2021randomized}. 

As far as we know, many past research endeavors focused on using the \emph{first-order} policy gradient information for DRL. Different from these works, in this paper, we are mainly interested in understanding the \emph{second-order Hessian} information and its role in training a policy network effectively and efficiently. Several pioneering research works have been reported lately to deepen our understanding of neural networks through the lens of the Hessian, primarily for the supervised learning paradigm \cite{yao2020pyhessian,dong2020hawq}. In the context of DRL, we found that different policy gradient algorithms can generate significantly different Hessian information (see our experiment results reported in Section \ref{sub-sec-perf-metric-tensor}). We hypothesize that properly utilizing and controlling such Hessian information can noticeably improve the performance of DRL algorithms.

One key Hessian information that attracted huge attention is the \emph{Hessian trace}. Referring to Section \ref{sec-prob}, minimizing the \emph{absolute Hessian trace} can be defined as an important regularization mechanism. In fact, the process of training a policy network can be conceived as an orbit in a high-dimensional \emph{policy parametric space}. Previous research either implicitly or explicitly treated this parametric space as an \emph{Euclidean-like manifold}, which is completely separated from the loss function \cite{martens2020new,zhang2019fast,kunstner2019limitations,chen2015using,chen2014stochastic,peng2020effective}. In other words, the \emph{metric tensor field} denoted as $g_{ab}$ on the manifold does not match the differential structure of the policy network and its loss function. Hence, the roughness of the loss function is translated directly to the roughness of the orbit, leading to compromised and unreliable learning performance.

We set the goal to develop new mathematical tools and DRL algorithms to learn a desirable metric tensor field $g_{ab}$ that transforms the policy parametric space into a \emph{generalized Riemannian manifold}. On such a manifold, policy training guided by its \emph{Levi-Civita connection} (aka. \emph{torsion-free $g_{ab}$ compatible derivative operator}) \cite{kreyszig2013differential} is expected to be smooth and reliable, resulting in improved effectiveness and sample efficiency. Motivated by this, we propose an essential criteria for the learned $g_{ab}$ to induce \emph{zero divergence} on the \emph{vector field} associated with the policy gradients. Zero divergence corresponds to the theoretical minimum of the absolute Hessian trace. It helps to nullify the principal differential components of a policy network and its loss function \cite{kampffmeyer2019deep,schafer2019vector,liu2023guo,chen2020learning}.

Notably, $g_{ab}$ is a complex geometric structure, learning which is beyond the capability of existing machine learning models \cite{roy2018geometry,le2015unsupervised,beik2021learning}. To make $g_{ab}$ regularized DRL feasible and effective, a new DNN architecture is deigned in this paper to significantly reduce the complexity involved in learning $g_{ab}$. Specifically, our \emph{metric tensor DNN} utilizes Fourier analysis techniques to reduce its parametric space \cite{rippel2015spectral}. A parametric matrix representation of high-dimensional special orthogonal groups \cite{gerken2021geometric,hutchinson2021lietransformer,chen2022hierarchical} is also developed to ease the training of the metric tensor DNN by exploiting the symmetries of $g_{ab}$.

The above development paves the way for designing a new \emph{$g_{ab}$ regularization algorithm}. The algorithm comprises two main components, namely (1) the component for learning the metric tensor DNN and (2) the component that uses the learned metric tensor DNN to compute $g_{ab}$ regularized policy gradients. It can be applied to a variety of policy gradient algorithms. We have specifically studied two state-of-the-art DRL algorithms, namely Soft Actor Critic (SAC) \cite{haarnoja2018} and Twin Delayed Deep Deterministic (TD3) \cite{fujimoto2018}. Experiments on multiple benchmark \emph{reinforcement learning} (RL) problems indicate that the new $g_{ab}$ regularization algorithm can effectively improve the performance and reliability of SAC and TD3.

{\bf \emph{Contributions}}: According to our knowledge, we are the first in literature to study mathematical and deep learning techniques to learn $g_{ab}$ and use $g_{ab}$ regularization algorithms to train policy networks. Our research extends the policy parametric space to a generalized Riemmanian manifold where critical differential geometric information about policy networks and DRL problems can be captured through the learned $g_{ab}$ and explicitly utilized to boost the learning performance.

\section{Related Works}
\label{sec-rw}

Many recent research works studied a variety of possible ways to estimate policy gradients for effective DRL. For example, Generalized Proximal Policy Optimization (GePPO) introduces a general clipping mechanism to support policy gradient estimation from off-policy samples, achieving a good balance between stability and sample efficiency \cite{queeney2021generalized}. Policy-extended Value Function Approximator (PeVFA) enhances conventional value function approximator by utilizing additional policy representations \cite{tang2022inputting}. This enhancement improves the accuracy of the estimated policy gradients. Efforts have also been made to control the \emph{bias} and \emph{variance} of the estimated policy gradients \cite{fujimoto2018,haarnoja2018, fan2021explaining,ZHANG202040}. For instance, clipped double Q-learning~\cite{fujimoto2018}, entropy regularization \cite{haarnoja2018}, action normalization \cite{wang2020striving_sop}, and Truncated
Quantile Critics (TQC) \cite{kuznetsov2020controlling} techniques have been developed to effectively reduce the estimation bias. All these research works assume that the policy parametric space adopts the Euclidean metric and is flat.

The development of natural policy gradient presents a major deviation from the flat parametric space \cite{liu2020improved,ding2020natural}. Its successful use on many challenging DRL problems clearly reveals the importance of expanding the policy parametric space to a generalized Riemannian manifold \cite{grondman2012survey}.
However, since the metric tensor field $g_{ab}$ for natural policy gradient is defined via the \emph{Fisher information matrix}, different from this paper, differential geometric information with regard to DRL problems is not utilized to learn $g_{ab}$ and boost learning performance.

We propose to learn $g_{ab}$ under the guidance of high-order Hessian information, particularly the Hessian trace, associated with the policy gradients. In the literature, notable efforts have been made towards understanding the influence of Hessian information on deep learning performance \cite{yao2020pyhessian,dong2020hawq,wu2020noisy,shen2019hessian,singla2019understanding}. For example, efficient numerical linear algebra (NLA) techniques have been developed in \cite{yao2020pyhessian} to compute top Hessian eigenvalues, Hessian trace, and Hessian eigenvalue spectral density of DNNs. In \cite{dong2020hawq}, Hessian trace is also exploited to determine suitable quantization scales for different layers of a DNN. Different from the past research works, instead of examining Hessian information in an Euclidean parametric space, we bring differential geometric techniques to alter and improve the differential structure of the parametric space. As far as we are aware, this is the first attempt towards achieving this goal within the existing body of literature.

\section{Background}
\label{sec-bc}

This paper considers the conventional DRL problems that can be modeled as \emph{Markov Decision Processes} (MDPs). Specifically, an MDP is a tuple $(\mathbb{S}, \mathbb{A}, P, R, \gamma)$, where $\mathbb{S}$ is the \emph{state space}, $\mathbb{A}$ is the \emph{action space}, $P$ stands for the state-transition probability function, $R$ is the reward function that provides immediate scalar feedback to a DRL agent, and $\gamma\in (0,1]$ is a \emph{discount factor}. More specifically, $P(s,a)$ captures the probability of transiting to any possible next state $s'\sim P(s,a)$ whenever the agent performs action $a\in\mathbb{A}$ in state $s\in\mathbb{S}$. In association with such state transition, a scalar reward is determined according to $R(s,a)$.

A \emph{policy} $\pi: \mathbb{S} \rightarrow \mathbb{A}$ produces an action $a\in\mathbb{A}$ (or a distribution over multiple actions) with respect to any state $s\in\mathbb{S}$. We can quantify the performance of any policy $\pi$ through a value function $v_{\pi}(s)$ below that predicts the \emph{expected discounted cumulative return} obtainable by following $\pi$ to interact with the learning environment, starting from $s\in \mathbb{S}$:
$$
v_{\pi}(s)=\mathop{\mathbb{E}}_{a_t \sim \pi}[\sum^{\infty}_{t=0}\gamma^{t}(R(s_{t},a_{t}) | s_0=s].
$$
Hence, an RL problem has the goal to find an \emph{optimal policy} $\pi^*$ that maximizes the value function with respect to any possible initial state $s_0 \in \mathbb{S}$. To make it feasible to solve large RL problems, the policy is often modeled as a parametric function in the form of a DNN. We denote such a parametric policy as $\pi_{\theta}$, where $\theta\in \mathbb{R}^n$ stands for the $n$-dimensional policy parameter, $n\gg 1$.

Starting from a randomly initialized policy parameter $\theta_0$, a policy network is repeatedly trained in the direction of its policy gradient defined below to gradually approach the optimal policy parameter, indicated as $\theta^*$:
$$
\nabla_{\theta} \mathbb{E}_{s_0}[v_{\pi_{\theta}}(s_0)]=\left[ \frac{\partial \mathbb{E}_{s_0}[v_{\pi_{\theta}}(s_0)]}{\partial \theta^{(0)}},\ldots, \frac{\partial \mathbb{E}_{s_0}[v_{\pi_{\theta}}(s_0)]}{\partial \theta^{(n)}} \right]^T
$$
where $\theta^{(i)}, 0\leq i\leq n$, denotes the $i$-th dimension of the policy parameter $\theta$. Estimating the policy gradient is at the core of many existing DRL algorithms and is the central focus of this paper. We will develop a $g_{ab}$ regularization method to approximate policy gradients in a generalized Riemannian manifold. The details of this regularization method is explained in the next two sections.

\section{Metric Tensor Regularized Policy Gradient}
\label{sec-prob}

In line with the introduction, we transform the $n$-dimensional policy parametric space to become a generalized Riemannian manifold $(\mathbb{R}^n, g_{ab})$, accompanied by a $(0,2)$-type metric tensor field $g_{ab}$ defined on $\mathbb{R}^n$ \cite{petersen2006riemannian}. Here we follow the \emph{abstract index notation} widely used in contemporary physics studies to represent tensors and their operations \cite{thorne2017modern}. Any $\theta\in \mathbb{R}^n$ comprises of $n$ trainable parameters in the policy network. Its tangent vector space on $\mathbb{R}^n$ is denoted as $T_{\theta}$. $g_{ab}$ satisfies two important properties with regard to $T_{\theta}$, $\forall\theta\in\mathbb{R}^n$:
\begin{equation*}
\begin{split}
(1) & \forall u^a,v^b\in T_{\theta}, g_{ab} u^a v^b = g_{ba} u^a v^b; \\
(2) & \text{Assume that } u^a \text{ satisfies the equation } g_{ab} u^a v^b =0, \forall v^b\in T_{\theta}, \text{then } u^a=0.
\end{split}
\end{equation*}
The first property above reveals the \emph{symmetric} nature of $g_{ab}$. The second property requires $g_{ab}$ to be \emph{non-degenerate}. Given any $g_{ab}$ that is $\mathrm{C}^{\infty}$ on $\mathbb{R}^n$, a torsion-free and $g_{ab}$ compatible derivative operator $\nabla_a$ can always be uniquely determined such that $\nabla_a g_{bc}=0$ on $\mathbb{R}^n$. Unless otherwise specified, $\nabla_a$ always refers to this compatible derivative operator in this paper. Using $\nabla_a$, the conventional policy gradient for $\forall \theta\in\mathbb{R}^n$ can be defined as a dual vector of $\theta$ below:
$$
\nabla_a \mathbb{E}_{s_0}[v_{\pi_{\theta}}(s_0)] = \partial_a \mathbb{E}_{s_0}[v_{\pi_{\theta}}(s_0)] = \sum_{\mu=1}^n \frac{\partial \mathbb{E}_{s_0}[v_{\pi_{\theta}}(s_0)]}{\partial \theta^{(\mu)}}  (\mathrm{d}\theta^{(\mu)})_a
$$
where $(\mathrm{d}\theta^{(\mu)})_a, 1\leq\mu\leq n,$ are the basis dual vectors of the dual vector space $T^*_{\theta}$ at $\theta\in\mathbb{R}^n$. $\partial_a$ is the so-called \emph{ordinary derivative operator}. Subsequently, the vector with respect to the conventional policy gradient at $\theta\in\mathbb{R}^n$ becomes:
$$
J^a|_{\theta}=g^{ab}\nabla_b \mathbb{E}_{s_0}[v_{\pi_{\theta}}(s_0)]=\sum_{\nu=1}^n \left(\sum_{\mu=1}^n g^{\nu,\mu} \frac{\partial \mathbb{E}_{s_0}[v_{\pi_{\theta}}(s_0)]} {\partial \theta^{(\mu)}} \right) \left( \frac{\partial}{\partial \theta^{(\nu)}} \right)^a
$$
in the manifold $(\mathbb{R}^n,g_{ab})$, where $(\partial/\partial \theta^{(\nu)})^a, 1\leq\nu\leq n,$ are the basis vectors of the vector space $T_{\theta}$ at $\theta$. We shall use $J^a|_{\theta}$ consistently as the \emph{vector representation} of the policy gradient in differential geometry. To obtain such a vector representation, we need to introduce the inverse metric tensor $g^{ab}$ that satisfies
$$
g^{ab} g_{bc} =\delta^{a}_{c} = \sum_{\mu=1}^n \sum_{\nu=1}^n\delta^{\mu}_{\nu} \left( \frac{\partial}{\partial \theta^{(\mu)}} \right)^a (\mathrm{d} \theta^{(\nu)})_c
$$
where $\delta^{a}_{c}$ above is the $(1,1)$-type \emph{identity tensor} such that $\delta^{a}_{b}v^b=v^a, \forall v^a\in T_{\theta},$ and $\delta^{a}_{b}w_a=w_b, \forall w_a\in T^*_{\theta}$. Accordingly, $\delta^{\mu}_{\nu}=1$ whenever $1\leq\mu=\nu\leq n$ and $\delta^{\mu}_{\nu}=0$ otherwise. In other words, if we represent $g_{ab}$ at any $\theta\in\mathbb{R}^n$ in the form of a matrix $G_{\theta}=[g_{\mu,\nu}(\theta)]_{\mu,\nu=1}^n$, then $g^{ab}$ can be determined as its inverse matrix $G_{\theta}^{-1}$. Hence the $g_{ab}$ \emph{regularized policy gradient} for training a policy network can be computed via a matrix expression below
\begin{equation}
\vec{J}|_{\theta} = G_{\theta}^{-1} \cdot \nabla_{\theta} \mathbb{E}_{s_0}[v_{\pi_{\theta}}(s_0)],
\label{equ-ja}
\end{equation}
which is a vector of real numbers. Such a vector is called a \emph{vector in linear algebra}. To distinguish it from a vector in differential geometry, we denote it in the above expression as $\vec{J}$ instead of $J^a$. Each real-valued dimension of $\vec{J}$ corresponds to a separate trainable parameter (or dimension) of the policy parametric space. The definition of $J^a|_{\theta}$ (and $\vec{J}|_{\theta}$) above allows us to construct a vector space in the manifold $(\mathbb{R}^n,g_{ab})$, indicated as $J^a$. 

\emph{Divergence} is a popular tool that captures important differential geometric structure of $J^a$. Specifically, based on $\nabla_a$, the divergence of $J^a$ can be mathematically defined as
$$
\forall\theta\in\mathbb{R}^n, Div (J^a)|_{\theta} = \nabla_a J^a|_{\theta}
$$
As a scalar quantity, $Div (J^a)|_{\theta}$ provides essential information about the distribution of the vectors on $(\mathbb{R}^n,g_{ab})$. Intuitively, if the vectors are moving away at any $\theta\in\mathbb{R}^n$, the divergence is positive, and if they are converging towards $\theta$, the divergence is negative. A \emph{zero divergence} indicates that the vectors are neither spreading nor converging at $\theta$. In the following, we demonstrate the potential advantages of achieving zero divergence everywhere in $(\mathbb{R}^n,g_{ab})$.

With respect to any $\theta\in\mathbb{R}^n$, we can perform a second-order Taylor expansion of $\mathbb{E}_{s_0}[v_{\pi_{\theta}}(s_0)]$ in the manifold $(\mathbb{R}^n,g_{ab})$, as presented below:
\begin{equation}
\mathbb{E}_{s_0}[v_{\pi_{(\theta+\vec{v})}}(s_0)] \approx \mathbb{E}_{s_0}[v_{\pi_{\theta}}(s_0)] + v^a \nabla_a \mathbb{E}_{s_0}[v_{\pi_{\theta}}(s_0)] + \frac{1}{2} v^a v^b \nabla_a \nabla_b \mathbb{E}_{s_0}[v_{\pi_{\theta}}(s_0)]
\label{equ-tay-exp}
\end{equation}
Here $v^a$ refers to an arbitrary vector at $\theta$ that can cause a small positional change of $\theta$ in the policy parametric space. We use $\vec{v}\in\mathbb{R}^n$ to indicate the same vector in classical linear algebra. Hence $\theta+\vec{v}$ corresponds to an different element of the manifold $(\mathbb{R}^n,g_{ab})$ that is close to $\theta$. Note that $v^a = g^{ab}v_b$, hence
$$
\frac{1}{2} v^a v^b \nabla_a \nabla_b \mathbb{E}_{s_0}[v_{\pi_{\theta}}(s_0)]=\frac{1}{2} v_b v^b g^{ab}\nabla_a \nabla_b \mathbb{E}_{s_0}[v_{\pi_{\theta}}(s_0)]
$$
If the divergence of $J^a$ is 0 at $\theta$, by definition $\nabla_b (g^{ab}\nabla_a \mathbb{E}_{s_0}[v_{\pi_{\theta}}(s_0)])|_{\theta}=0$. Therefore
$$
\nabla_b (g^{ab}\nabla_a \mathbb{E}_{s_0}[v_{\pi_{\theta}}(s_0)])|_{\theta}=g^{ab}\nabla_a\nabla_b \mathbb{E}_{s_0}[v_{\pi_{\theta}}(s_0)])|_{\theta}=0.
$$
In other words, by jointly considering all $n$ dimensions of $\theta$, we have
$$
\sum_{\mu=1}^n \sum_{\nu=1}^n g^{\mu,\nu} \nabla_{\theta^{(\mu)}}\nabla_{\theta^{(\nu)}} \mathbb{E}_{s_0}[v_{\pi_{\theta}}(s_0)]) = 0.
$$
Assume that $v^a$ satisfies $v^{(\mu)} v_{(\mu)}=c^2, \forall \mu=1,\ldots,n$, where $c$ is a real constant. $v^{(\mu)}$ and $v_{(\mu)}$ refer respectively to the $\mu$-th dimension of vector $v^a$ and its corresponding dual vector $v_a$ at $\theta$. Using this condition, we can simplify the Taylor expansion in \eqref{equ-tay-exp} as:
\begin{equation*}
\begin{split}
\mathbb{E}_{s_0}[v_{\pi_{(\theta+\vec{v})}}(s_0)] & \approx \mathbb{E}_{s_0}[v_{\pi_{\theta}}(s_0)] + \sum_{\mu=1}^n v^{(\mu)} \frac{\partial \mathbb{E}_{s_0}[v_{\pi_{\theta}}(s_0)]}{\partial \theta^{(\mu)}} + \frac{1}{2} \sum_{\mu=1}^n \sum_{\nu=1}^n v^{\nu} v_{\nu} g^{\mu,\nu} \nabla_{\theta^{(\mu)}}\nabla_{\theta^{(\nu)}} \mathbb{E}_{s_0}[v_{\pi_{\theta}}(s_0)]) \\
& = \mathbb{E}_{s_0}[v_{\pi_{\theta}}(s_0)] + \sum_{\mu=1}^n v^{(\mu)} \frac{\partial \mathbb{E}_{s_0}[v_{\pi_{\theta}}(s_0)]}{\partial \theta^{(\mu)}} + \frac{c^2}{2} \sum_{\mu=1}^n \sum_{\nu=1}^n g^{\mu,\nu} \nabla_{\theta^{(\mu)}}\nabla_{\theta^{(\nu)}} \mathbb{E}_{s_0}[v_{\pi_{\theta}}(s_0)]) \\
& = \mathbb{E}_{s_0}[v_{\pi_{\theta}}(s_0)] + \sum_{\mu=1}^n v^{(\mu)} \frac{\partial \mathbb{E}_{s_0}[v_{\pi_{\theta}}(s_0)]}{\partial \theta^{(\mu)}}.
\end{split}
\end{equation*}
Therefore, with zero divergence, the second-order differential components involved in approximating $\mathbb{E}_{s_0}[v_{\pi_{(\theta+\vec{v})}}(s_0)]$ can be nullified. Since the above approximation guides the training of policy networks in practice, we believe $g_{ab}$ regularized policy gradient in \eqref{equ-ja} can improve the reliability and performance of policy gradient based DRL algorithms. Driven by this motivation, we aim to develop mathematical tools and deep learning techniques to achieve $g_{ab}$ regularized policy network training in the next section.

\section{Metric Tensor Regularization Method for Training Policy Networks}
\label{sec-alg}

A DRL algorithm can use $g_{ab}$ regularized policy gradient in \eqref{equ-ja} to train the policy parameters $\theta$ of a policy network $\pi_{\theta}$. Such an algorithm has the goal to find the optimal policy parameters $\theta^*$, the same as many existing DRL algorithms \cite{schulman2017proximal,lillicrap2015,schulman2015}. As mentioned in Section \ref{sec-int}, our $g_{ab}$ regularization method comprises of two components, which will be introduced respectively in Subsections \ref{subsec-learn-gab} and \ref{subsec-use-gab}. As a generally applicable machine learning technique, we will further apply the $g_{ab}$ regularization method to SAC and TD3 to develop practically useful DRL algorithms in Subsection \ref{subsec-alg-gab}.

\subsection{Learning a DNN Model of $g_{ab}$}
\label{subsec-learn-gab}

As a $(0,2)$-type symmetric tensor on $\mathbb{R}^n$, $g_{ab}$ at any $\theta\in\mathbb{R}^n$ can be represented as an $n\times n$ symmetric matrix $G_{\theta}=[g_{\mu,\nu}(\theta)]_{\mu,\nu=1}^n$ where each row and column of $G_{\theta}$ correspond to one dimension of the policy parametric space. Meanwhile, each element of matrix $G_{\theta}$, i.e. $g_{\mu,\nu}(\theta)$, is a function of $\theta$. Learning such a matrix representation of $g_{ab}$ directly is a challenging task, since $n\gg 1$ for most of policy networks used in DRL algorithms. To make it feasible to learn $g_{ab}$ in the form of a DNN, we impose a specific structure on $G_{\theta}$, as given below:
\begin{equation}
G_{\theta}=I_n + \vec{u}(\theta)\cdot \vec{u}(\theta)^T
\label{equ-G}
\end{equation}
where $I_n$ stands for the $n\times n$ identity matrix. $\vec{u}(\theta): \mathbb{R}^n \rightarrow \mathbb{R}^n$ is a vector-valued function of $\theta$. Hence $\vec{u}(\theta)\cdot\vec{u}(\theta)^T$ produces an $n\times n$ matrix. It is easy to verify that the simplified matrix $G_{\theta}$ in \eqref{equ-G} is symmetric and non-degenerate. Hence it is suitable to serve as the matrix representation of $g_{ab}$. According to Section \ref{sec-prob}, we aim to learn $g_{ab}$ that can induce zero divergence on the vector field $J^a$ in manifold $(\mathbb{R}^n, g_{ab})$. Driven by this objective and following \eqref{equ-G}, we can obtain Proposition \ref{prop-1} below to compute the divergence of $J^a$ at any $\theta\in\mathbb{R}^n$ efficiently. A proof of Proposition \ref{prop-1} can be found in Appendix A.

\begin{proposition}
Given a metric tensor field $g_{ab}$ with its matrix representation defined in \eqref{equ-G} in manifold $(\mathbb{R}^n, g_{ab})$, the divergence of $\mathrm{C}^{\infty}$ vector field $J^a$ at any $\theta\in\mathbb{R}^n$, i.e. $Div(J^a)|_{\theta}$, is
$$
Div(J^a)|_{\theta}=\sum_{\mu=1}^n \left( \frac{\partial \vec{J}^{(\mu)}}{\partial \theta^{(\mu)}} + \frac{\vec{J}^{(\mu)}}{1+\vec{u}(\theta)^T\cdot \vec{u}(\theta)}\sum_{\nu=1}^n \vec{u}^{(\nu)}(\theta)\frac{\partial \vec{u}^{(\nu)}(\theta)}{\partial \theta^{(\mu)}} \right)
$$
where $\vec{J}^{(\mu)}$ refers to the $\mu$-th component of vector $J^a|_{\theta}$ at $\theta$, which corresponds to the $\mu$-th dimension of the policy parametric space. Similarly, $\theta^{(\mu)}$ and $\vec{u}^{(\nu)}$ represent the $\mu$-th component of $\theta$ and $\nu$-th component of vector $\vec{u}(\theta)$, respectively.
\label{prop-1}
\end{proposition}

In theory, $\vec{u}(\theta)$ in \eqref{equ-G} can be arbitrary functions of $\theta$. To tackle the complexity of learning $\vec{u}(\theta)$, we can re-formulate $\vec{u}(\theta)$ in the form of a parameterized linear transformation of $\theta$, i.e.
$$
\vec{u}(\theta,\phi) = T(\theta,\phi) \cdot \theta
$$
where $T(\theta,\phi)$ is an $n\times n$ matrix associated with $\theta$ and parameterized by $\phi$, $dim(\phi)=m$ and $m\ll n$. $\theta$ is treated as an $n$-dimensional vector in the above matrix expression. We can further simplify the linear transformation introduced via $T(\theta,\phi)$ into a combination of two elementary operations, i.e. \emph{rotation} and \emph{scaling}. Subsequently, we can re-write $T(\theta,\phi)$ as
\begin{equation}
T(\theta,\phi)=S(\theta,\phi_1) \cdot R(\theta,\phi_2)
\label{equ-T-decom}
\end{equation}
where $S(\theta,\phi_1)$ stands for the $n\times n$ \emph{scaling matrix} parameterized by $\phi_1$, $R(\theta,\phi_2)$ stands for the $n\times n$ \emph{rotation matrix} parameterized by $\phi_2$. The combination of $\phi_1$ and $\phi_2$ gives rise to $\phi$. Hence, $dim(\phi_1)+dim(\phi_2)=dim(\phi)=m$.

The scaling matrix $S(\theta,\phi_1)$ controls the magnitude of each dimension of $\vec{u}(\theta)$. We can specifically represent $S(\theta,\phi_1)$ as a diagonal matrix, i.e. $S(\theta,\phi_1)=Diag(\vec{\omega}(\theta,\phi_1))$. The diagonal line of matrix $S(\theta,\phi_1)$ forms an $n$-dimensional vector $\vec{\omega}(\theta,\phi_1)$. While it sounds straightforward to let $\omega(\theta,\phi_1)=\phi_1$, this implies that $dim(\phi_1)=n$, contradicting with the requirement that $m\ll n$. To tackle this issue, we perform Fourier transformation of $\vec{\omega}$ and only keep the low-frequency components of $\vec{\omega}$ which can be further controlled via $\phi_1$. Specifically, define a series of $n$-dimensional vectors $\vec{\Omega}^{(i)}$ using the trigonometrical function $cos()$ as
$$
\vec{\Omega}^{(i)}=\sqrt{\frac{2}{{n}}} \left[
\begin{array}{c}
cos\left( \frac{2\pi i}{n} j \right)|_{j=0} \\
\vdots \\
cos\left( \frac{2\pi i}{n} j \right)|_{j=n-1}
\end{array}
\right],
$$
where $1\leq i\leq \tilde{m}$ and $\tilde{m}< m$. Further define $\Omega$ as an $n\times \tilde{m}$ matrix:
$$
\Omega = [\vec{\Omega}^{(1)}, \ldots, \vec{\Omega}^{(\tilde{m})}]
$$
Then $\vec{\omega}(\theta,\phi_1)$ can be obtained through the matrix expression below:
\begin{equation}
\vec{\omega}(\theta,\phi_1)=\Omega \cdot \vec{\tilde{w}}(\theta,\phi_1),
\label{equ-vec-omega}
\end{equation}
with $\vec{\tilde{w}}(\theta,\phi_1)$ being an $\tilde{m}$-dimensional vector parameterized by $\phi_1$ that controls the magnitude of low-frequency components of $\vec{\omega}$. Consequently, the problem of learning the $n\times n$ scaling matrix $S(\theta,\phi_1)$ is reduced to the problem of learning the vector $\vec{\tilde{\omega}}(\theta,\phi_1)$ with $\tilde{m}\ll n$ dimensions.

Compared to $S(\theta,\phi_1)$, it is more sophisticated to learn the parameterized rotation matrix $R(\theta,\phi_2)$. In group theory, any $n\times n$ rotation matrix serves as the matrix representation of a specific element of the $n$-dimensional \emph{Special Orthogonal} (SO) group, denoted as $SO(n)$ \cite{hall2013lie}. Consider the Lie algebra of $SO(n)$, indicated as $\mathcal{SO}(n)$. $\mathcal{SO}(n)$ is defined mathematically below
$$
\mathcal{SO}(n)=\{ n\times n \text{ real-valued matrix } A | A^T=-A \}.
$$
In other words, $\mathcal{SO}(n)$ is the set of all $n\times n$ \emph{anti-symmetric matrices}. Consequently, $\exp(A)$ must be an $n\times n$ rotation matrix, $\forall A\in\mathcal{SO}(n)$. In view of this, a straightforward approach is to construct $R(\theta,\phi_2)$ directly from $A$. However, because $dim(\mathcal{SO}(n))=\frac{n(n-1)}{2}$, we cannot treat all independent elements of $A$ as $\phi_2$ in $R(\theta,\phi_2)$, since $dim(\mathcal{SO}(n))> n\gg m$. To simplify the parameterization of $R(\theta,\phi_2)$, we introduce Proposition \ref{prop-2} below. Its proof can be found in Appendix B.

\begin{proposition}
Assume that $A\in\mathcal{SO}(n)$, there exist $n\times n$ unitary matrices $U$ and $V$ such that
$$
\exp(A)=U\cdot\Sigma_c\cdot U^T-V\cdot\Sigma_s\cdot U^T
$$
where, with respect to an $n$-dimensional vector $\vec{\sigma}=[\sigma^{(1)},\ldots,\sigma^{(n)}]^T$, $\Sigma_c$ and $\Sigma_s$ are defined respectively as
$$
\Sigma_c=\left[
\begin{array}{ccc}
cos(\sigma^{(1)}) & \ & 0 \\
\ & \ddots & \ \\
0 & \ & cos(\sigma^{(n)})
\end{array}
\right]
\mathrm{\ and\ }
\Sigma_s=\left[
\begin{array}{ccc}
sin(\sigma^{(1)}) & \ & 0 \\
\ & \ddots & \ \\
0 & \ & sin(\sigma^{(n)})
\end{array}
\right]
$$
\label{prop-2}
\end{proposition}

Following Proposition \ref{prop-2}, we can simplify the construction of $R(\theta,\phi_2)$. Notice that
$$
(\vec{\Omega}^{(i)})^T\cdot\vec{\Omega}^{(j)}\approx\left\{
\begin{array}{ll}
1, & i=j \\
0, & i\neq j
\end{array}
\right. , \forall i,j\in\{1,\ldots,\tilde{m} \}
$$
$\Omega$ can be utilized to approximate the first unitary matrix $U$ in Proposition \ref{prop-2}. Similarly, we can define another series of $n$-dimensional vectors $\vec{\Phi}^{(i)}$ as
$$
\vec{\Phi}^{(i)}=\sqrt{\frac{2}{{n}}} \left[
\begin{array}{c}
sin\left( \frac{2\pi i}{n} j \right)|_{j=0} \\
\vdots \\
sin\left( \frac{2\pi i}{n} j \right)|_{j=n-1}
\end{array}
\right],
$$
where $1\leq i\leq \tilde{m}$. $\Phi=[\vec{\Phi}^{(1)},\ldots,\vec{\Phi}^{(\tilde{m})}]$ gives a good approximation of the second unitary matrix $V$ in Proposition \ref{prop-2}. However, different from $V$ and $U$, which are $n\times n$ matrices, $\Omega$ and $\Phi$ are $n\times \tilde{m}$ matrices. To cope with this difference in dimensionality, we introduce a parameterized $\tilde{m}$-dimensional vector $\vec{\tilde{\sigma}}(\theta,\phi_2)$. Assume that functions $cos()$ and $sin()$ are applied elementary-wise to $\vec{\tilde{\sigma}}(\theta,\phi_2)$, then
$$
\tilde{\Sigma}_c=Diag(cos(\vec{\tilde{\sigma}}(\theta,\phi_2))) \text{ and } \tilde{\Sigma}_s=Diag(sin(\vec{\tilde{\sigma}}(\theta,\phi_2)))
$$
are $\tilde{m}\times\tilde{m}$ diagonal matrices. Subsequently, define
\begin{equation}
\tilde{R}(\theta,\phi_2)= \Omega\cdot\tilde{\Sigma}_c\cdot\Omega^T-\Phi\tilde{\Sigma}_s\cdot\Omega^T.
\label{equ-rm-approx}
\end{equation}
Similar to the Fourier transformation technique applied to the scaling matrix, \eqref{equ-rm-approx} also draws inspiration from frequency analysis, as clearly revealed by Proposition \ref{prop-3} below. The proof of this proposition can be found in Appendix C.

\begin{proposition}
For any $A\in\mathcal{SO}(n)$, assume that $\exp(A)=\hat{\Omega}\cdot\Sigma_c\cdot\hat{\Omega}^T-\hat{\Phi}\cdot\Sigma_s\cdot\hat{\Omega}^T$, where $\hat{\Omega}$ and $\hat{\Phi}$ are defined similarly as $\Omega$ and $\Phi$ with the additional requirement that $\tilde{m}=n$. Hence $\hat{\Omega}$ and $\hat{\Phi}$ are $n\times n$ unitary matrices. Under this assumption, for any $n$-dimensional vector $\vec{a}$,
$$
\exp(A)\cdot \vec{a}=\sum_{i=1}^n \eta_i \sqrt{\frac{2}{n}} \left[
\begin{array}{c}
cos\left( \frac{2\pi i}{n}j+\vec{\sigma}^{(i)}|_{j=0} \right) \\
\vdots \\
cos\left( \frac{2\pi i}{n}j+\vec{\sigma}^{(i)}|_{j=n-1} \right)
\end{array}
\right]
$$
where $\eta_i=(\vec{\Omega}^{(i)})^T\cdot \vec{a}$ stands for the magnitude of the $i$-th frequency component of $\vec{a}$\footnote{Vector $\vec{a}$ in Proposition \ref{prop-3} is treated as a signal. The temporal index of this signal corresponds to each dimension of $\vec{a}$.}.
\label{prop-3}
\end{proposition}

Proposition \ref{prop-3} indicates that, upon applying a rotation matrix $\exp(A)$ to any vector $\vec{a}$, this will lead to independent phase shift of each frequency component of $\vec{a}$, controlled by the respective dimension of vector $\vec{\sigma}$. Hence, $\tilde{R}(\theta,\phi_2)$ in \eqref{equ-rm-approx} only shifts/rotates the low frequency components (i.e. the first $\tilde{m}$ low frequency components) of a vector. In view of this, a complete parametrized rotation matrix can be constructed as
\begin{equation}
R(\theta,\phi_2)=\tilde{R}(\theta,\phi_2)+I_n-\Omega\cdot\Omega^T.
\label{equ-rm-approx-c}
\end{equation}
Whenever we multiply $R(\theta,\phi_2)$ in \eqref{equ-rm-approx-c} with any vector $\vec{a}$, only the low-frequency components of vector $\vec{a}$ is phase shifted or rotated. The high-frequency components of $\vec{a}$ remain untouched. As a result, the problem of learning the $n\times n$ rotation matrix $R(\theta,\phi_2)$ is reduced to the problem of learning the $\tilde{m}$-dimensional vector $\vec{\tilde{\sigma}}(\theta,\phi_2)$ parameterized by $\phi_2$.

\eqref{equ-T-decom}, \eqref{equ-vec-omega}, \eqref{equ-rm-approx} and \eqref{equ-rm-approx-c} together give rise to a complete parameterized model for $\vec{u}(\theta,\phi)$ and subsequently for $G_{\theta}$ in \eqref{equ-G}. Given \eqref{equ-G} and hence $g_{ab}$, the divergence of the vector space $J^a$ at $\theta\in\mathbb{R}^n$ can be calculated according to Proposition \ref{prop-1}. Therefore, the problem for learning $g_{ab}$ can be formulated below as an optimization problem:
\begin{equation}
\min_{\phi} (Div(J^a)|_{\theta})^2 =\min_{\phi} \left(
\sum_{\mu=1}^n \left( \frac{\partial \vec{J}^{(\mu)}}{\partial \theta^{(\mu)}} + \frac{\vec{J}^{(\mu)}}{1+\vec{u}(\theta,\phi)^T\cdot \vec{u}(\theta,\phi)}\sum_{\nu=1}^n \vec{u}^{(\nu)}(\theta,\phi)\frac{\partial \vec{u}^{(\nu)}(\theta,\phi)}{\partial \theta^{(\mu)}} \right)
\right)^2
\label{equ-gab-learn}
\end{equation}
By finding $\phi$ that minimizes the square of the divergence above, we expect to bring the actual divergence of $J^a$ close to 0 at any $\theta\in\mathbb{R}^n$. Note that, in practice, we don't need to consider all possible policy parameters in the manifold $(\mathbb{R}^n,g_{ab})$. Instead, whenever we obtain any policy parameter $\theta$ during DRL, we can update $\phi$ towards minimizing $(Div(J^a)|_{\theta})^2$. In other words, the learning of $g_{ab}$ can be localized during the training of the policy network $\pi_{\theta}$. In fact, we can build a DNN that processes $\theta$ as its input and produces $\vec{\tilde{\omega}}(\theta,\phi_1)$ and $\vec{\tilde{\sigma}}(\theta,\phi_2)$ as its output. Therefore, $\phi_1$ and $\phi_2$ are the trainable parameters of this DNN, which will be called the \emph{metric tensor DNN}. More details about its architecture design are presented in Section \ref{sec-exp}.

\subsection{Using the Learned $g_{ab}$ Model to Compute Regularized Policy Gradient}
\label{subsec-use-gab}

Based on the metric tensor DNN as a deep model of $g_{ab}$, in this subsection, we will develop two alternative methods to compute $g_{ab}$ regularized policy gradient. The {\bf first method} directly follows \eqref{equ-ja}. Specifically, according to the Sherman-Morrison formula \cite{press2007numerical},
$$
G_{\theta}^{-1}=I_n-\frac{\vec{u}(\theta,\phi)\cdot \vec{u}(\theta,\phi)^T}{1+\vec{u}(\theta,\phi)^T\cdot\vec{u}(\theta,\phi))}
$$
Consequently,
\begin{equation}
\vec{J}|_{\theta}=\nabla_{\theta} \mathbb{E}_{s_0}[v_{\pi_{\theta}}(s_0)] - \frac{ \vec{u}(\theta,\phi)^T\cdot \nabla_{\theta} \mathbb{E}_{s_0}[v_{\pi_{\theta}}(s_0)]}{1+\vec{u}(\theta,\phi)^T\cdot\vec{u}(\theta,\phi))} \vec{u}(\theta,\phi)
\label{equ-gab-reg-gradient-compute}
\end{equation}

The {\bf second method} enables us to update $\theta$ along the direction of a \emph{geodesic} \cite{kreyszig2013differential}, which is jointly and uniquely determined by the metric tensor in manifold $(\mathbb{R}^n,g_{ab})$ and the policy gradient vector. Geodesic is a generalization of straight lines in manifold $(\mathbb{R}^n,g_{ab})$. Many existing optimization algorithms follow geodesics to search for optimal solutions in high-dimensional manifolds \cite{hu2020brief}. For simplicity and clarity, we use the term \emph{geodesic regularized policy gradient} to indicate the direction of the geodesic that passes through the policy parameter $\theta$ in manifold $(\mathbb{R}^n,g_{ab})$. Note that geodesic regularized policy gradient depends on $g_{ab}$ and hence should be viewed as $g_{ab}$ regularized policy gradient too. However, in order to clearly distinguish it from \eqref{equ-gab-reg-gradient-compute}, we give it a different name. Proposition \ref{prop-4} below provides an efficient way to estimate geodesic regularized policy gradient. Its proof can be found in Appendix D.

\begin{proposition}
Given the manifold $(\mathbb{R}^n,g_{ab})$ for the policy parametric space, at any $\theta\in\mathbb{R}^n$, a geodesic $\Gamma$ that passes through $\theta$ can be uniquely and jointly determined by $g_{ab}$ and the $g_{ab}$ regularized policy gradient vector $J^a|_{\theta}$ at $\theta$. Assume that $g_{ab}$ changes smoothly and stably along $\Gamma$, there exist $\zeta_1,\zeta_2>0$ such that the geodesic regularized policy gradient at $\theta$ can be approximated as
$$
\vec{T}^{(\delta)}|_{\theta}\approx\vec{J}^{(\delta)}|_{\theta} + \zeta_1 \vec{J}^{(\delta)}|_{\theta} + \zeta_2 \sum_{\rho=1}^n g^{\delta\rho}(\theta)\sum_{\mu=1}^n \sum_{\nu=1}^n \frac{\partial g_{\mu,\nu}(\theta)}{\partial \theta^{(\rho)}} J^{(\mu)}|_{\theta} J^{(\nu)}|_{\theta}
$$
where $\vec{T}^{(\delta)}|_{\theta}$ stands for the $\delta$-th dimension of the geodesic regularized policy gradient $\vec{T}$ at $\theta$, $0\leq\delta\leq n$.
\label{prop-4}
\end{proposition}

It is straightforward to see from Proposition \ref{prop-4} that the geodesic regularized policy gradient is closely related to the $g_{ab}$ regularized policy gradient $\vec{J}|_{\theta}$, which is further adjusted by an additional term controlled by a positive coefficient $\zeta_2$. In practice, we can treat $\frac{\zeta_2}{1+\zeta_1}$ as a hyper-parameter of our $g_{ab}$ regularization algorithm. 

\subsection{DRL algorithms based on $g_{ab}$ Regularized Policy Gradient}
\label{subsec-alg-gab}

Following the mathematical and algorithmic developments in Subsections \ref{subsec-learn-gab} and \ref{subsec-use-gab}, a new $g_{ab}$ regularization algorithm is created, as presented in the pseudo-code form in Algorithm \ref{alg-gab-reg-alg}.

\begin{algorithm}[htb!]
\caption{The Metric Tensor Regularization Algorithm} \label{alg-gab-reg-alg}
\begin{algorithmic}
\item Based on the up-to-date $\theta$, compute conventional policy gradient $\nabla_{\theta} \mathbb{E}_{s_0}[v_{\pi_{\theta}}(s_0)]$;
\item Using the metric tensor DNN, compute $J^a|_{\theta}$ and $Div(J^a)|_{\theta}$ using \eqref{equ-gab-reg-gradient-compute} and Proposition \ref{prop-1}; 
  \While{the maximum number of iterations has not been reached}
    \State Update $\phi$ of the metric tensor DNN towards minimizing $(Div(J^a)|_{\theta})^2$;
    \State Re-compute $J^a|_{\theta}$ and $Div(J^a)|_{\theta}$.
  \EndWhile
\item Re-compute $J^a|_{\theta}$ based on the trained metric tensor DNN;
\item Compute geodesic regularized policy gradient $\vec{T}|_{\theta}$ using Proposition \ref{prop-4};
\item Return $J^a|_{\theta}$ and $\vec{T}|_{\theta}$.
\end{algorithmic}
\end{algorithm}

Algorithm \ref{alg-gab-reg-alg} starts from the calculation of the conventional policy gradient $\nabla_{\theta} \mathbb{E}_{s_0}[v_{\pi_{\theta}}(s_0)]$ with respect to the most recently learned policy parameter $\theta$. This can be realized by using various existing DRL algorithms such as SAC and TD3. Afterwards, based on the metric tensor DNN, we compute the $g_{ab}$ regularized policy gradient vector $J^a$ as well as its divergence at $\theta$ by using \eqref{equ-gab-reg-gradient-compute} and Proposition \ref{prop-1} respectively. Guided by the square of the computed divergence as the loss function, Algorithm \ref{alg-gab-reg-alg} updates trainable parameters $\phi$ of the metric tensor DNN towards achieving near-zero divergence at $\theta$\footnote{We set the maximum number of training iterations in Algorithm \ref{alg-gab-reg-alg} to 20 in the experiments. We can further increase this number but it does not seem to produce any noticeable performance gains.}. Based on the trained metric tensor DNN, the $g_{ab}$ regularized policy gradient as well as the corresponding geodesic regularized policy gradient will be computed by Algorithm \ref{alg-gab-reg-alg} as its output. These gradients will be further utilized to train policy network $\pi_{\theta}$.

Building on Algorithm \ref{alg-gab-reg-alg}, we can modify existing DRL algorithms to construct their $g_{ab}$ regularized counterparts. We have specifically considered two state-of-the-art DRL algorithms, namely SAC and TD3, due to their widespread popularity in literature. However, the practical application of Algorithm \ref{alg-gab-reg-alg} is not restricted to SAC and TD3. It remains as an important future work to study the effective use of Algorithm \ref{alg-gab-reg-alg} in other DRL algorithms.

\begin{algorithm}[htb!]
\caption{The Metric Tensor Regularized Policy Gradient Algorithm} \label{alg-gab-pol-grad}
\begin{algorithmic}
\item Initialize policy network $\pi_{\theta}$ with randomly sampled $\theta\in\mathbb{R}^n$;
\For{each sampled episode till the maximum number of episodes is reached}
    \State Store all sampled state transitions into the replay buffer;
    \State Randomly sample a mini-batch from the replay buffer;
    \State Compute conventional policy gradient by using SAC, TD3 or other policy gradient algorithms;
    \State Compute $g_{ab}$ regularized and geodesic regularized policy gradients by using Algorithm \ref{alg-gab-reg-alg};
    \State Train policy network $\pi_{\theta}$ by using regularized policy gradients;
\EndFor
\State Return the trained policy network $\pi_{\theta}$.
\end{algorithmic}
\end{algorithm}

Algorithm \ref{alg-gab-pol-grad} gives a high-level description of $g_{ab}$ regularized DRL algorithms. Algorithm \ref{alg-gab-pol-grad} is designed to be compatible with SAC, TD3 and many other policy gradient algorithms. Following Algorithm \ref{alg-gab-pol-grad}, we can identify four main algorithm variations, including SAC that uses $g_{ab}$ regularized policy gradients, SAC that uses geodesic regularized policy gradients, TD3 that uses $g_{ab}$ regularized policy gradients, and TD3 that uses geodesic regularized policy gradients. We call these variations respectively as SAC-J, SAC-T, TD3-J, and TD3-T. All these variations will be experimentally examined in Section \ref{sec-exp}.

\section{Experiments}
\label{sec-exp}

\subsection{Experiment Setting}

We use the popular OpenAI Spinning Up repository~\cite{SpinningUp2018} to implement $g_{ab}$ regularized DRL algorithms introduced in the previous section. Our implementation follows closely all hyper-parameter setting and network architectures reported in~\cite{haarnoja2018,fujimoto2018} and summarized in Table~\ref{tab:hyper-para}. Since calculating the Hessian trace precisely can pose significant computation burden on existing deep learning libraries such as PyTorch, we adopt a popular Python library named PyHessian~\cite{yao2020pyhessian}, where Hutchinson's method~\cite{avron2011randomized,bai1996some} is employed to estimate the Hessian trace efficiently. All experiments were conducted on a cluster of Linux computing nodes with 2.5 GHz Intel Core
i7 11700 processors and 16 GB memory. To ensure consistency, all experiments were run in a virtual environment with Python 3.7.11 managed by the Anaconda platform. The main Python packages used in our experiments are summarized in Table~\ref{tab:python-lib}.

\begin{table}[htb!]
\caption{Hyper-parameter settings of all experimented algorithms.}
\label{tab:hyper-para}
\centering
\resizebox{0.9\textwidth}{!}{
\begin{tabular}{l||lll||lll}
\hline
Hyper-parameter       & SAC & SAC-J & SAC-T & TD3 & TD3-J & TD3-T  \\ \hline
Total training timesteps  & 300,000  & 300,000  & 300,000  & 300,000  & 300,000 & 300,000 \\
Max episode length        & 1000  & 1000 & 1000  & 1000  & 1000  & 1000 \\
Minibatch size        & 256 & 256 & 256 & 100 & 100 & 100   \\
Adam learning rate    & 3e-4 & 3e-4 & 3e-4  & 1e-3 & 1e-3  & 1e-3  \\
Discount ($\gamma$)          & 0.99  & 0.99 & 0.99  & 0.99 & 0.99 & 0.99 \\
GAE parameter ($\lambda$)     & 0.995 & 0.995 & 0.995 & 0.995  & 0.995  & 0.995 \\
Replay buffer size    & 1e6 & 1e6 & 1e6 & 1e6 & 1e6 & 1e6 \\
Update interval (timesteps)      & 50  & 50 & 50  & 50  & 50  & 50 \\
Network architecture & 256x256 & 256x256 & 256x256 & 400x300 &400x300 & 400x300 \\
\hline
\end{tabular}}
\end{table}

\begin{table}[htb!]
\caption{Python packages.}
\label{tab:python-lib}
\centering
\begin{tabular}{l||l}
\hline
Package name      &Version    \\ \hline
cython    &0.29.25    \\
gym       &0.21.0     \\
mujoco-py   &2.1.2.14    \\
numpy    &1.21.4   \\
pybulletgym       &0.1 \\
python    &3.7.11       \\
PyHessian &0.1\\
torch    &1.13.1 \\
\hline
\end{tabular}
\end{table}

\begin{figure*}[tb]
    \centering
    \includegraphics[width=1\linewidth]{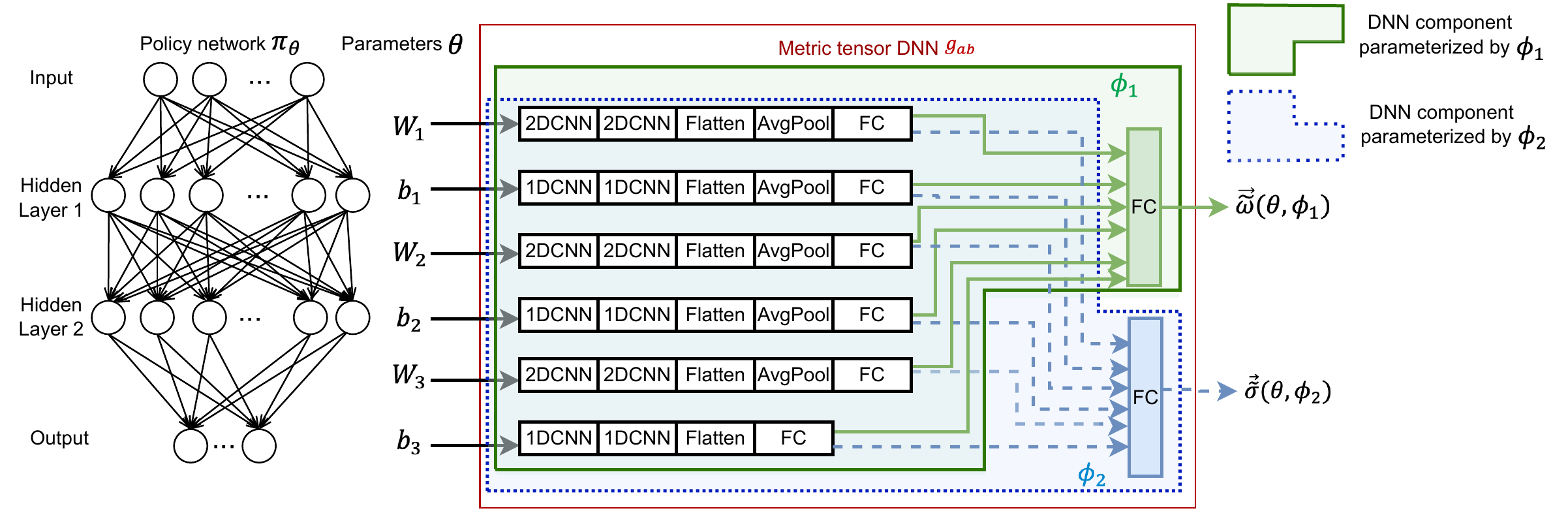}
    \caption{The metric tensor DNN designed to learn the metric tensor field $g_{ab}$ of the policy parametric space. The metric tensor DNN processes a policy network $\pi_{\theta}$ as its input and produces $\vec{\tilde{\omega}}(\theta,\phi_1)$ and $\vec{\tilde{\sigma}}(\theta,\phi_2)$ as its output. $\phi_1$ and $\phi_2$ together define the trainable parameters of the metric tensor DNN. Specifically, we denote the parameters contained within the green hexagon as $\phi_1$ and those within the blue hexagon as $\phi_2$. The intersection of these two hexagons corresponds to the common parameters shared between $\phi_1$ and $\phi_2$ in the metric tensor DNN. $W_i$ and $b_i$ refer respectively to the weight matrix and the bias vector of the $i$-th layer of the policy network $\pi_{\theta}$.}
    \label{fig:gab_network}
\end{figure*}

To learn the complex geometric structure of $g_{ab}$, we introduce a new DNN architecture. This is exemplified by the \emph{metric tensor DNN} for a policy network $\pi_\theta$ with two hidden layers, as depicted in Figure~\ref{fig:gab_network}. The metric tensor DNN parameterized by $\phi_1$ and $\phi_2$ maps the $n$-dimensional policy parameter $\theta = [W_1, b_1, ..., W_3, b_3]$ into two $\tilde{m}$-dimensional vectors $\vec{\tilde{\omega}}(\theta,\phi_1)$ and $\vec{\tilde{\sigma}}(\theta,\phi_2)$, which are used to build the scaling matrix $S(\theta,\phi_1)$ and the rotation matrix $R(\theta,\phi_2)$ in \eqref{equ-T-decom} respectively\footnote{On LunnarLanderContinuous-v2, the input dimension $n$ of the metric tensor DNN is 69124 for a policy with two hidden layers described in Table~\ref{tab:hyper-para} for SAC-T. The output dimension is 700, effectively yielding two $\tilde{m}$-dimensional vectors ($\tilde{m}$=350).}. 

In particular, each layer of weight matrix $W_i$ and bias vector $b_i$ of $\pi_\theta$ is processed individually through two consecutive convolutional kernels (2D kernels of size 3$\times$3 for processing $W_i$ or 1D kernels of size 3 for processing $b_i$), followed by the flattening and average pooling operations with a pool size of 5 before passing through a dense layer. It should be noted that the bias vector for the output layer is exempt from the pooling operation due to its comparatively low dimensionality\footnote{Note that the dimension of the bias vector for the output layer of the policy network $\pi_{\theta}$ equals to the dimensionality of the action space, which is usually small. For example, the dimension is 2 for the LunnarLanderContinuous-v2 problem.}. The \emph{Softplus} function serves as the activation mechanism for both the convolutions and dense layers in the metric tensor DNN. We also used the ReLU activation function and obtained similar experiment results. The outputs of the dense layer are concatenated and then channeled through two dense layers, each of which yields an $\tilde{m}$-dimensional vector.

While the performance of the metric tensor DNN could be further enhanced through fine-tuning the DNN architecture, such an undertaking is beyond the scope of this paper. Consequently, we reserve the exploration of more advanced network architectural designs and fine-tuning for our future work. Moreover, the performance of the learned metric tensor DNN reported in Section~\ref{sub-sec-perf-metric-tensor} shows that our proposed simple architecture for the metric tensor DNN can effectively learn $g_{ab}|_{\theta}$ with respect to any policy parameter $\theta$ such that the divergence at $\theta$ can be made closer to 0.

Experiments are conducted on multiple challenging continuous control benchmark problems provided by OpenAI gym~\cite{openai_gym} (e.g., Hopper-v3, LunarLanderContinuous-v2, Walker2D-v3) and pyBullet~\cite{benelot2018} (e.g., Hopper-v0, Walker2D-v0 ). Each benchmark problem has a fixed maximum episode length of 1,000 timesteps. Each DRL algorithm is trained for $3\times 10^5$ timesteps. To obtain the cumulative returns, we average the results of 10 independent testing episodes after every 1,000 training timesteps for each individual algorithm run. Every competing algorithm was also run for 10 independent times to determine its average performance, which is reported in the following subsection.

\subsection{Experiment Result}

\subsubsection{Performance Comparison}\label{sub-sec-perf-comp}

The performance comparison between SAC and its metric tensor regularized variations, SAC-J and SAC-T, is presented in Table~\ref{tab:sac_final_perf_comp}. As the table clearly indicates, SAC-T significantly outperforms SAC on all benchmark problems. SAC-T also outperforms SAC-J on most of the benchmark problems, except the LunnarLanderContinuous-v2 problem, where SAC-T achieved 89\% of the highest cumulative returns obtained by SAC-J. Furthermore, in the case of the Hopper-v3 problem, SAC-T achieved over 50\% higher cumulative returns in comparison to SAC and 25\% higher cumulative returns when compared to SAC-J. Meanwhile, we found that using $g_{ab}$ regularized policy gradient alone may not frequently lead to noticeable performance gains since SAC-J performed better than SAC on two benchmark problems but also performed worse on one benchmark problem. These results suggest that policy parameter training should follow the direction of the geodesics in the Rimannian manifold $(\mathbb{R}^n,g_{ab})$ in order for $g_{ab}$ regularized policy gradient to effectively improve the performance of DRL algorithms. This observation agrees well with existing optimization techniques in Rimennian manifolds \cite{hu2020brief}.

\begin{table*}[!ht]
\caption{Final performance of all competing algorithms on 4 benchmark problems. }\label{tab:sac_final_perf_comp}
\centering
\resizebox{.8\textwidth}{!}{
\begin{tabular}{c||ccc}
\hline
Benchmark problems                       & SAC            & SAC-J            & SAC-T        \\ \hline 
Hopper-v3 (Mujoco)             & 2202.47$\pm$660.32   &  2714.03$\pm$559.77  & \textbf{3399.7$\pm$52.1}   \\
LunarLanderContinuous-v2         & 199.04$\pm$120.66  &  \textbf{245.42$\pm$0.0}  & 217.87$\pm$53.55 \\
Walker2D-v3 (Mujoco)             & 1689.15$\pm$786.91 &  1290.35$\pm$0.0  &  \textbf{2127.67$\pm$342.16}  \\ 
Hopper-v0 (PyBullet)             & 1575.66$\pm$396.78  & 1489.94$\pm$675.02  & \textbf{1968.81$\pm$212.94}   \\ \hline
\end{tabular}}
\end{table*}

\begin{figure*}[hbt]
    \begin{center}
        \subfloat[Hopper-v3 (Mujoco)]{\label{fig-hopper}\includegraphics[width=0.45\linewidth]{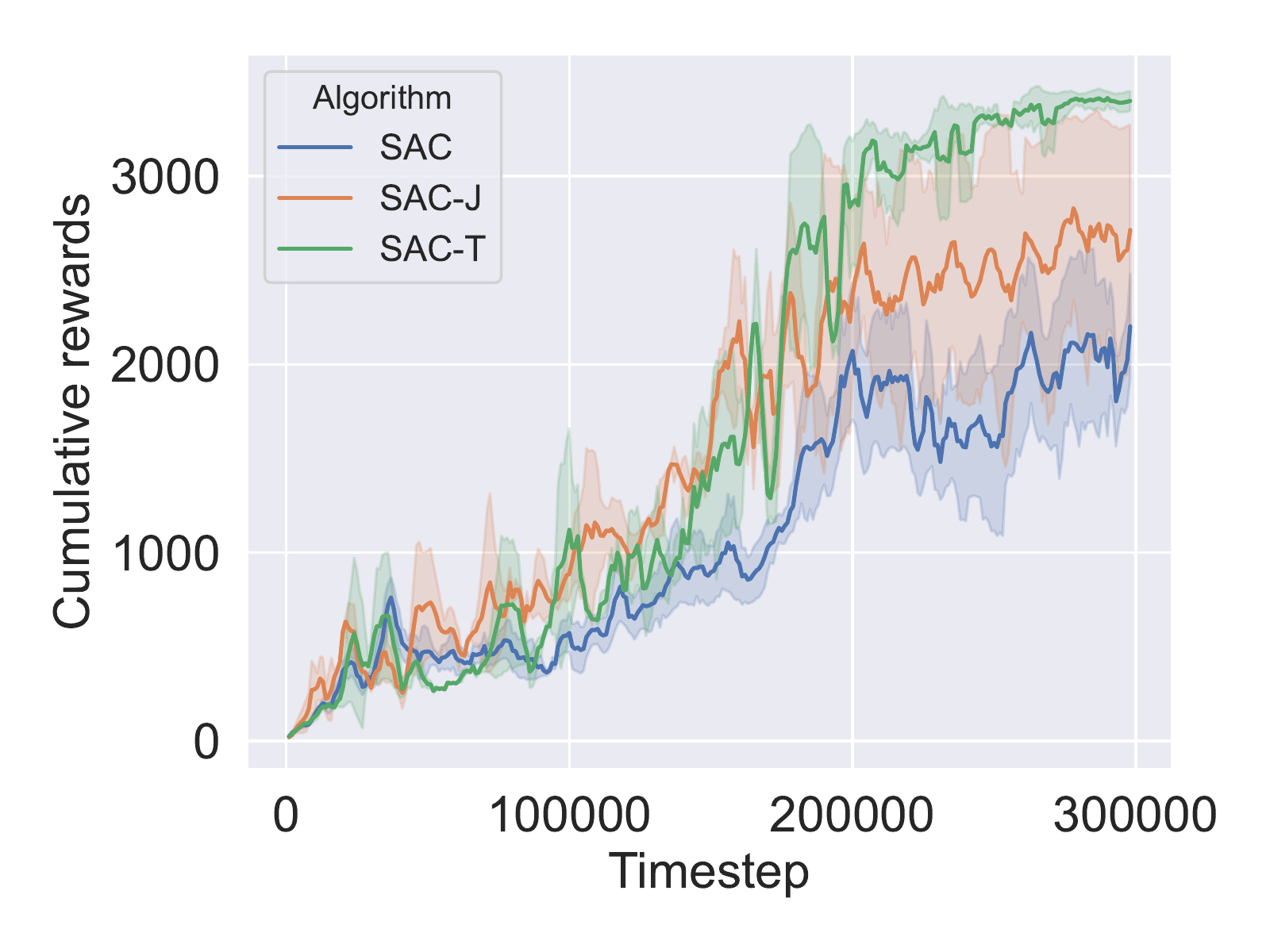}}
        \subfloat[LunarLanderContinuous-v2]{\label{fig-ll}\includegraphics[width=0.45\linewidth]{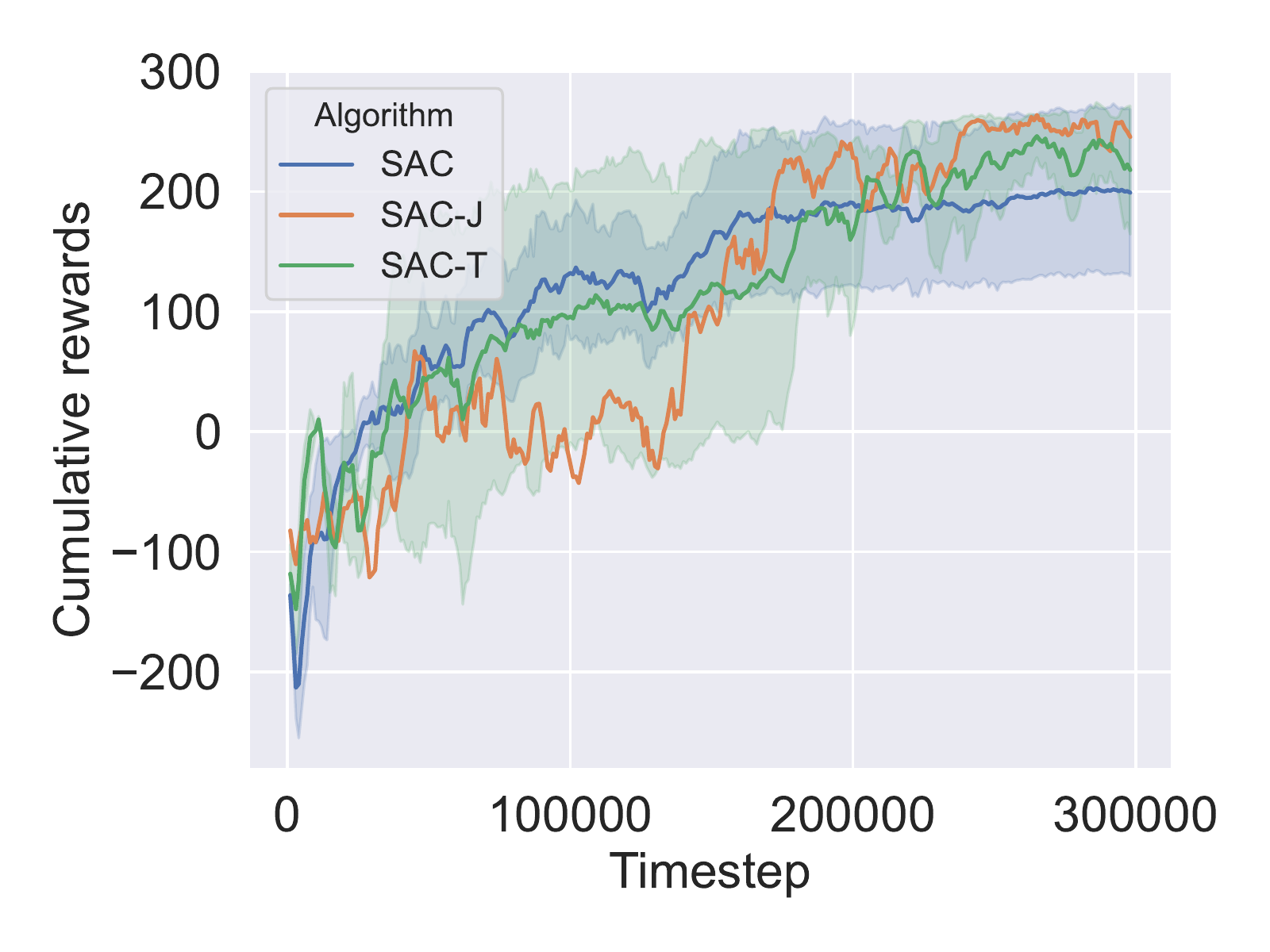}}\\
        \subfloat[Walker2D-v3 (Mujoco)]{\label{fig-walker}\includegraphics[width=0.45\linewidth]{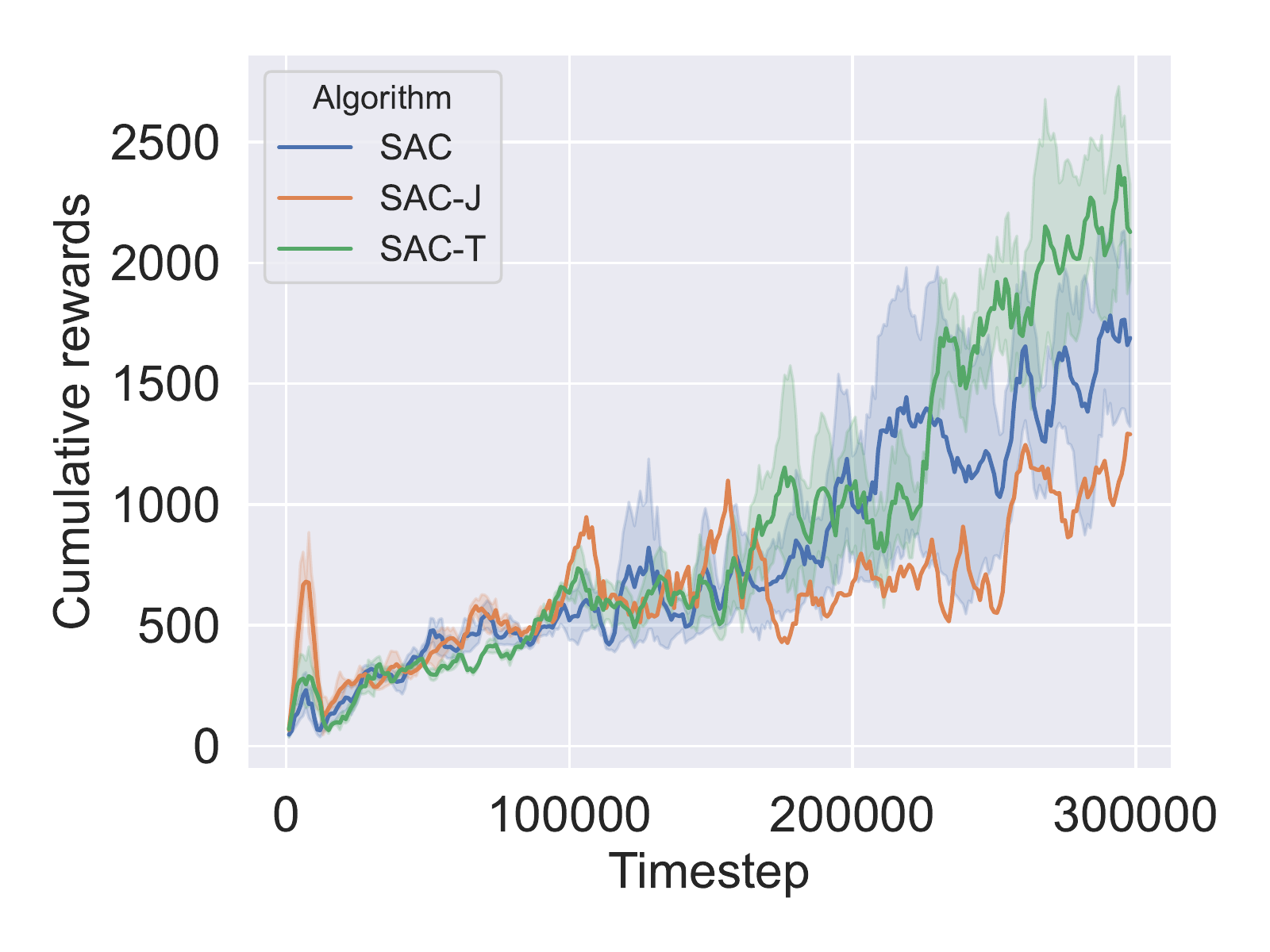}}
        \subfloat[Hopper-v0 (PyBullet)]{\label{fig-hopperPB}\includegraphics[width=0.45\linewidth]{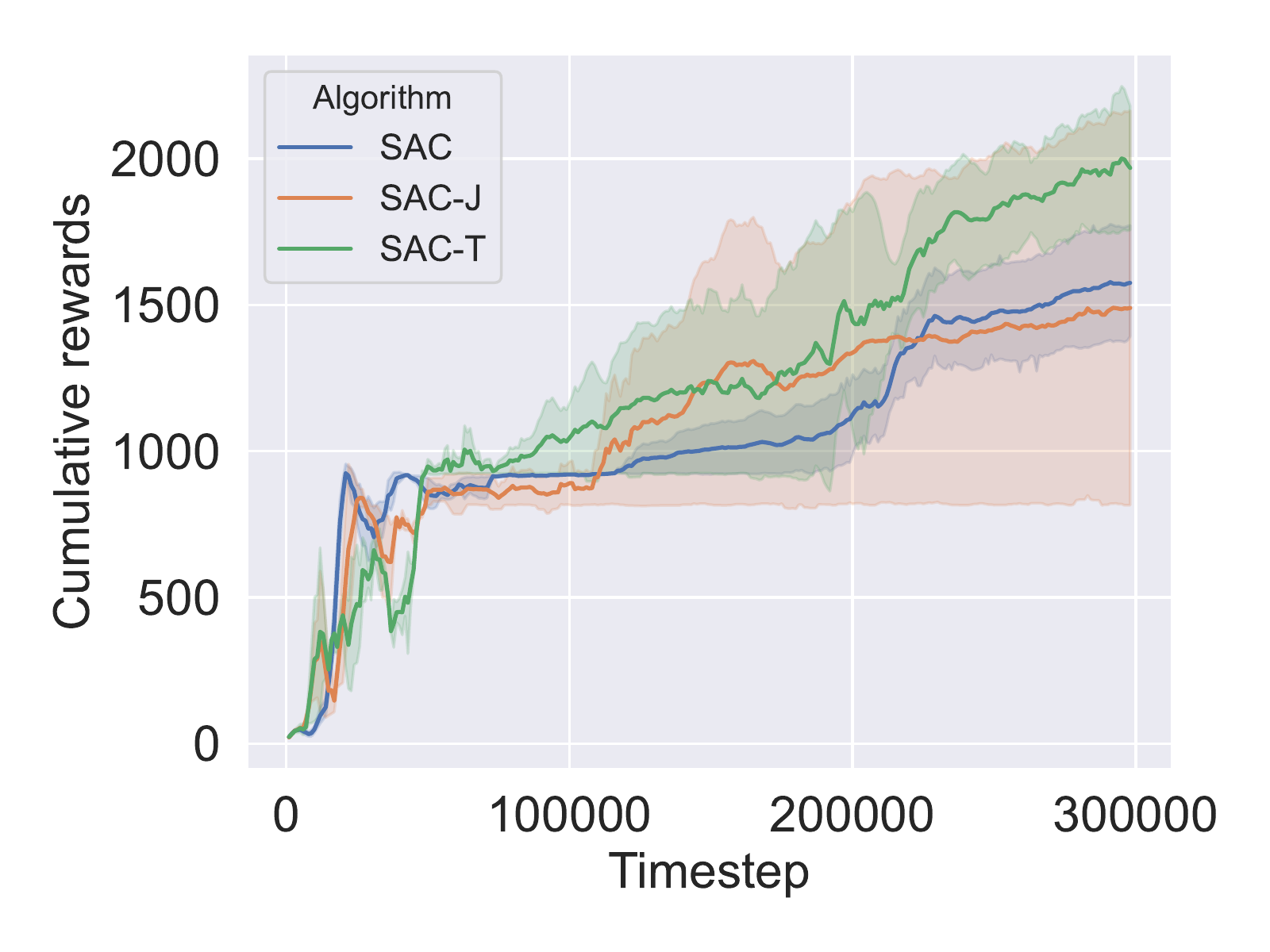}}
    \end{center}
    \caption{Learning curves of SAC, SAC-J, and SAC-T on four benchmark RL problems.}
    \label{fig:sac_training_perf}
\end{figure*}

The learning curve comparison of all competing algorithms is depicted in Figure~\ref{fig:sac_training_perf}. By transforming the policy parametric space into a generalized Riemannian manifold and guiding the policy parameter update along the geodesics in the manifold, SAC-T exhibits better stability during the learning process as compared to SAC and SAC-J. This increased stability is particularly noticeable on Hopper-v3 and Walker2D-v3, where SAC-T demonstrated reduced variations in comparison to other competing algorithms. 

Additionally, we compare the learning of TD3, TD3-J, and TD3-T on two benchmark problems. As demonstrated in Table~\ref{tab:td3_final_perf_comp}, we notice that the potential performance gains achievable by using $g_{ab}$ regularized policy gradients in TD3 is not as prominent as in SAC. Meanwhile, it is worthwhile to note that $g_{ab}$ regularized policy gradients will not weaken the performance of TD3. In fact, TD3-T demonstrated highly competitive performance, compared to TD3. This observation not only supports our previous findings but also demonstrates the broad applicability of our proposed metric tensor regularization algorithm.


\begin{table*}[!ht]
\caption{Final performance of all competing algorithms on two benchmark problems. }
\label{tab:td3_final_perf_comp}
\centering
\resizebox{.8\textwidth}{!}{
\begin{tabular}{c||ccc}
\hline
Benchmark problems                       & TD3            & TD3-J            & TD3-T        \\ \hline 
LunarLanderContinuous-v2         &  \textbf{276.98$\pm$4.38}  & 268.24$\pm$2.37 & 275.53$\pm$2.12 \\
Walker2D-v0 (PyBullet)             & 1327.33$\pm$206.0 &  1364.34$\pm$272.45  &  \textbf{1550.95$\pm$190.5}\\ \hline
\end{tabular}}
\end{table*}

\subsubsection{Further analysis of the metric tensor learning technique}\label{sub-sec-perf-metric-tensor}

\begin{figure*}[!hbt]
    \begin{center}
        \subfloat[Hopper-v3 (Mujoco)]{\label{fig-hopper}\includegraphics[width=0.45\linewidth]{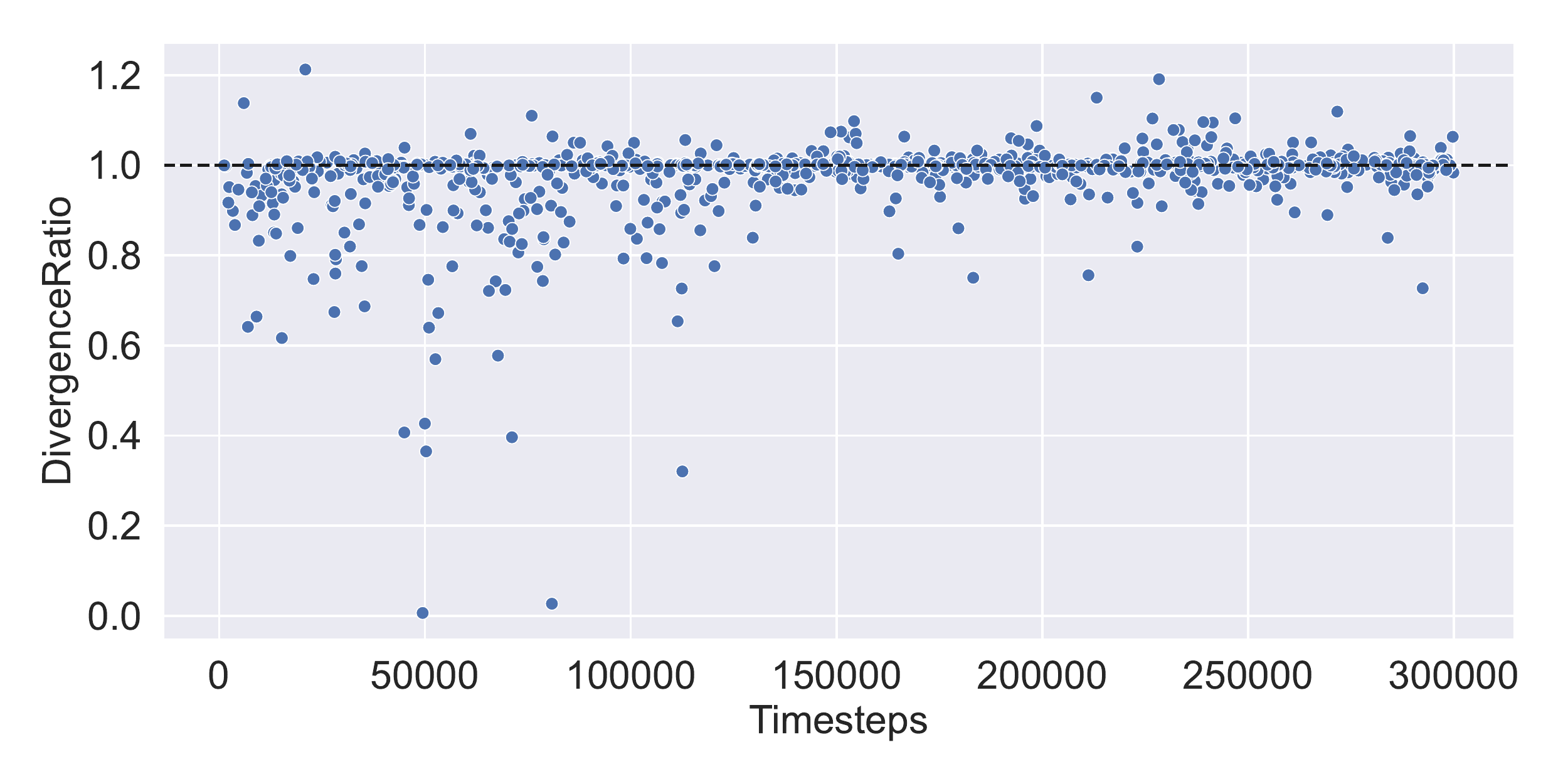}}
        \subfloat[LunarLanderContinuous-v2]{\label{fig-ll}\includegraphics[width=0.48\linewidth]{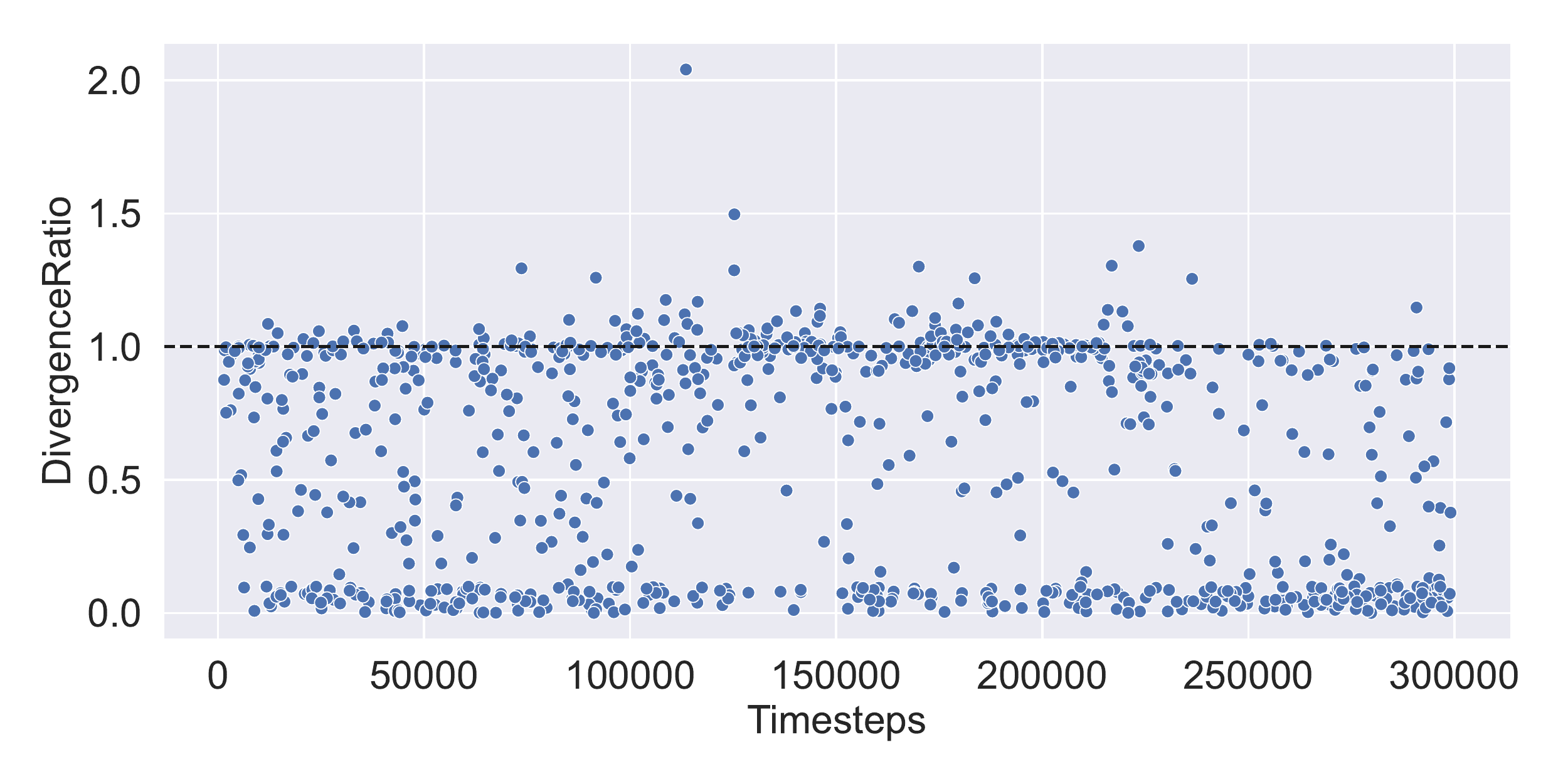}}\\
        \subfloat[Walker2D-v3 (Mujoco)]{\label{fig-walker}\includegraphics[width=0.45\linewidth]{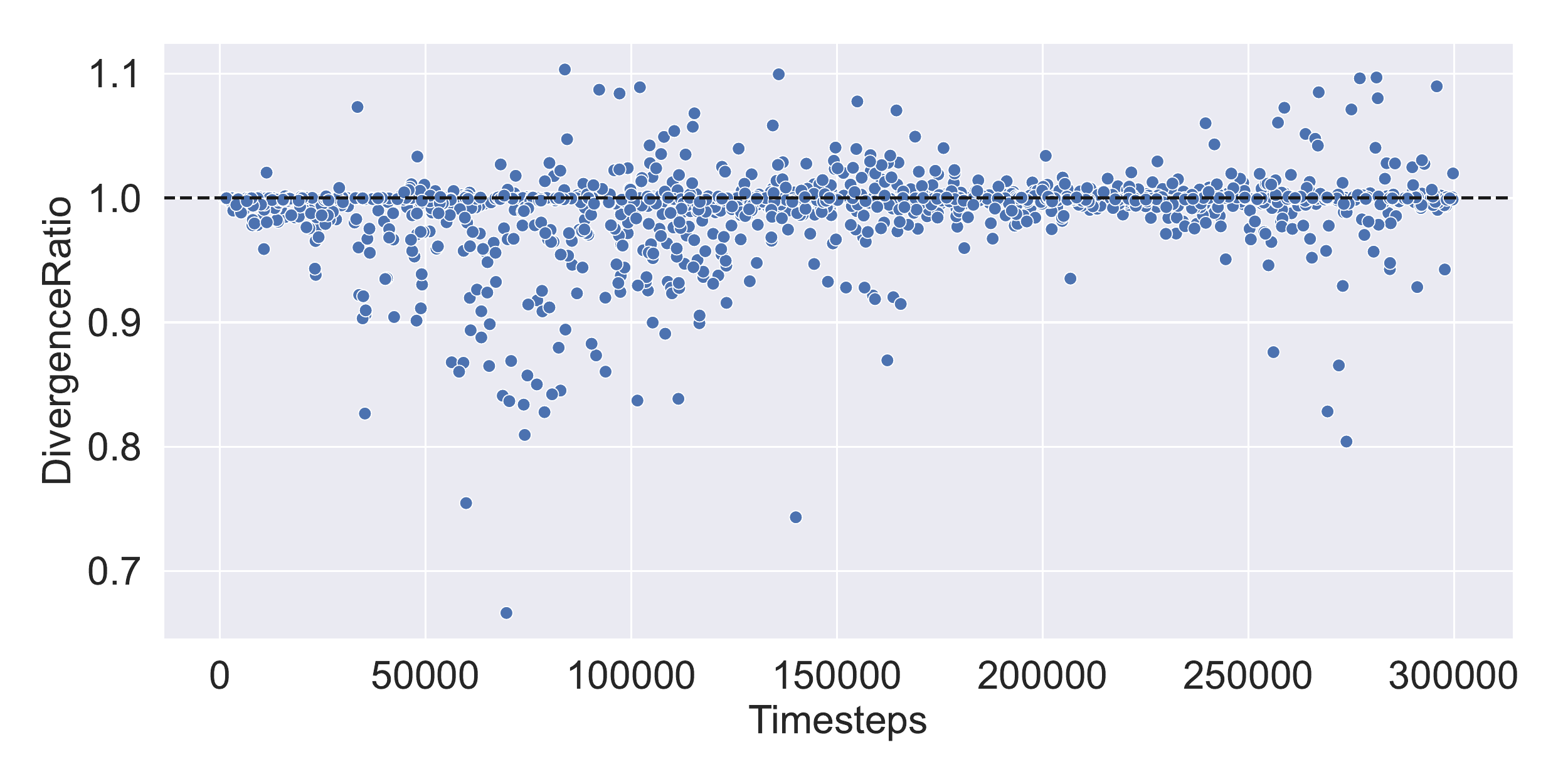}}
        \subfloat[Hopper-v0 (PyBullet)]{\label{fig-hopperPB}\includegraphics[width=0.48\linewidth]{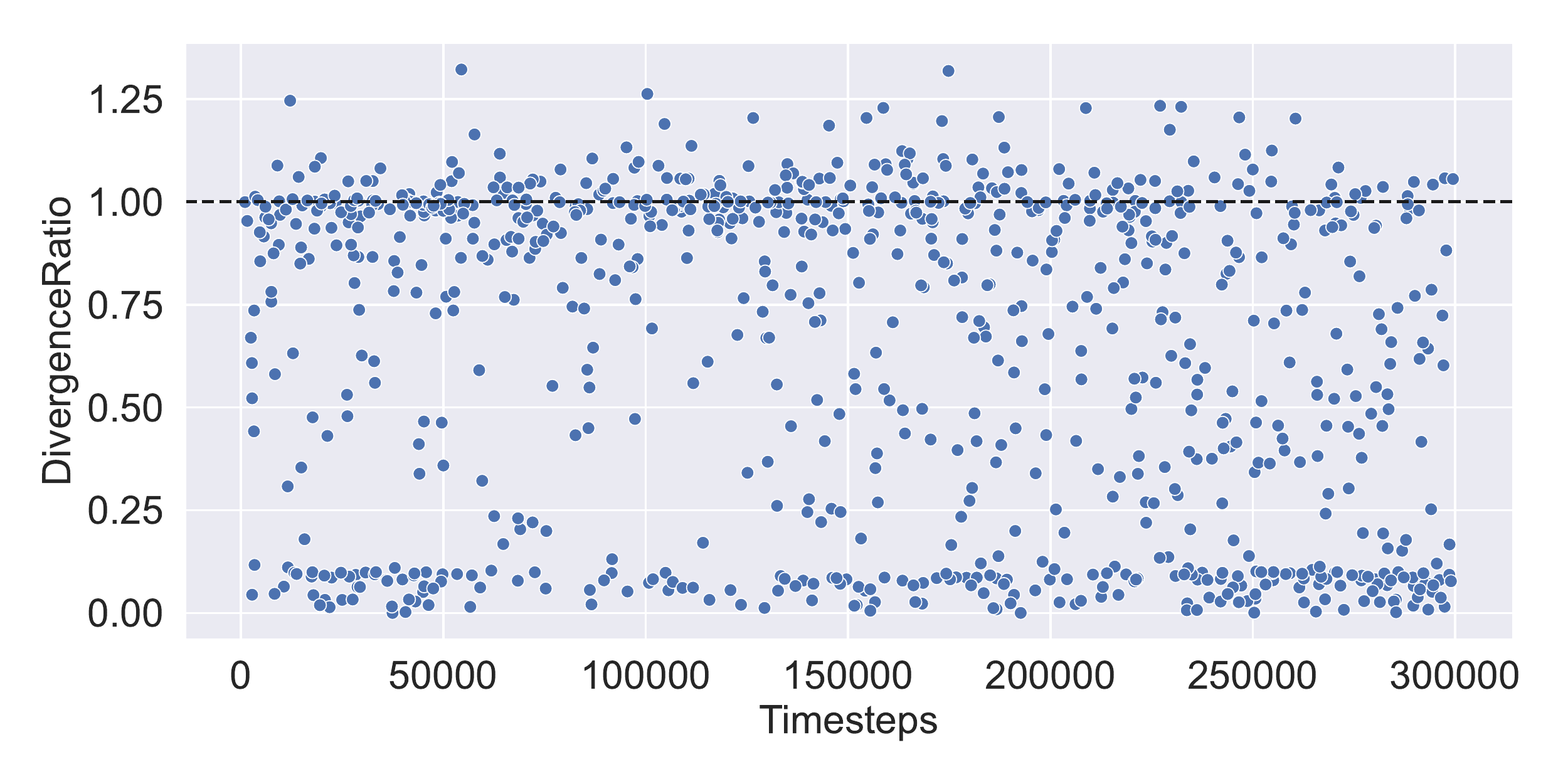}}
    \end{center}
    \caption{Divergence ratio of SAC-T during the training process, where the divergence ratio is defined as the absolute ratio between $Div(J^a)|_\theta$ and the Hessian trace.}
    \label{fig:sac_divergencesac_training_perf}
\end{figure*}

\begin{figure*}[!hbt]
    \begin{center}
        \subfloat[LunarLanderContinuous-v2]{\label{fig-ll}\includegraphics[width=0.48\linewidth]{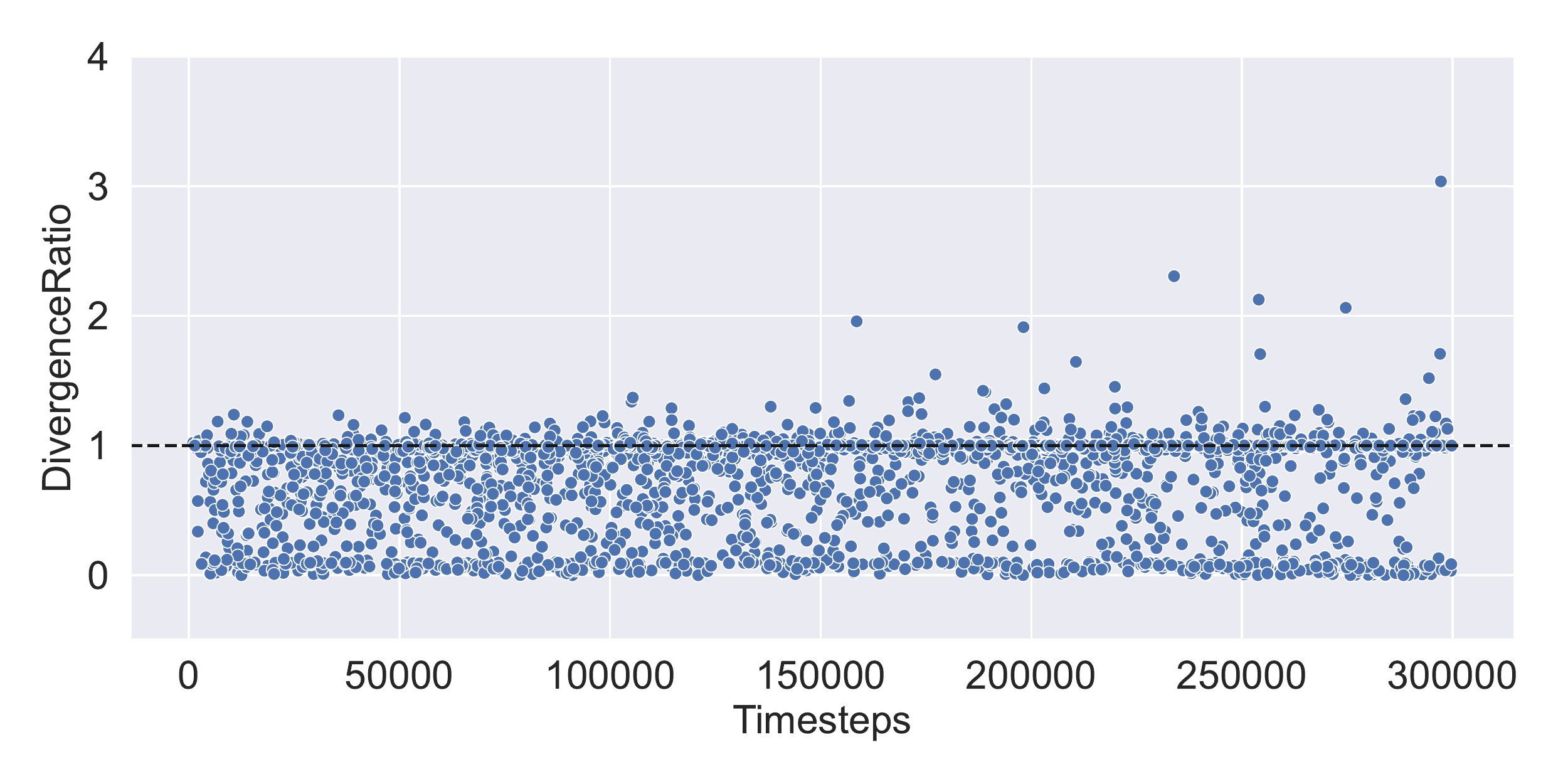}}
        \subfloat[Walker2D-v0 (PyBullet)]{\label{fig-walker}\includegraphics[width=0.45\linewidth]{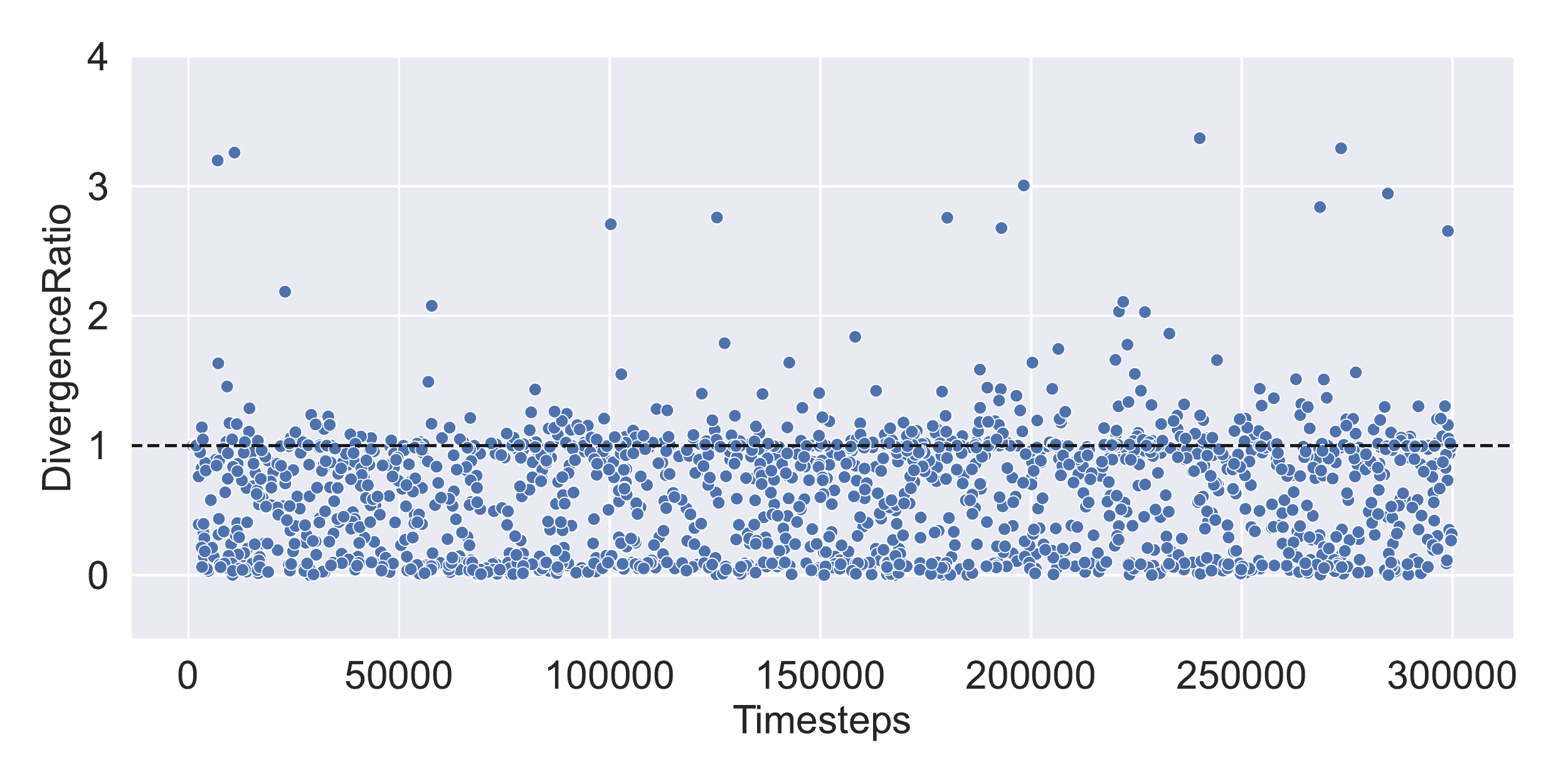}}
    \end{center}
    \caption{Divergence ratio of TD3-T during the training process. The percentages of Divergence ratios<1 for TD3-T are 74.56\% on LunnarLanderContinuous-v2 and 81.04\% on Walker2D-v0.}
    \label{fig:td3_divergencesac_training_perf}
\end{figure*}

\begin{table}[htb!]
\caption{The percentage of Divergence ratio < 1 for SAC-T on four benchmark problems.}
\label{tab:divergence-ratio}
\centering
\begin{tabular}{l||l}
\hline
Benchmark problems      & Divergence ratio < 1 (\%)  \\ \hline
Hopper-v3 (Mujoco)    &62.46    \\
LunarLanderContinuous-v2      &84.70     \\
Walker2D-v3 (Mujoco)  &64.52    \\
Hopper-v0 (PyBullet)    &79.58  \\
\hline
\end{tabular}
\end{table}

\begin{figure*}[!hbt]
    \begin{center}
        \subfloat[Hopper-v3 (Mujoco)]{\label{fig-hopper}\includegraphics[width=0.45\linewidth]{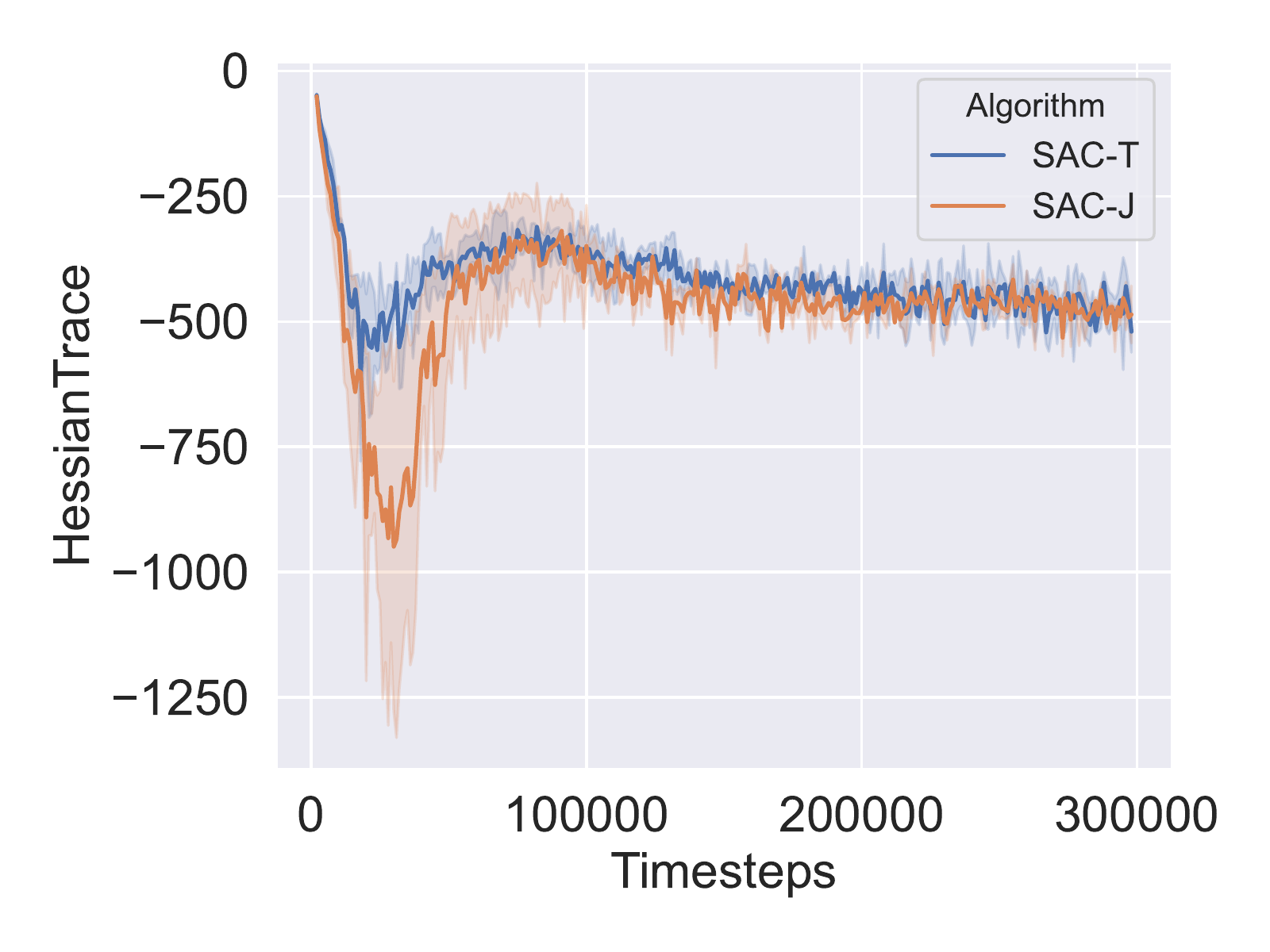}}
        \subfloat[LunarLanderContinuous-v2]{\label{fig-ll}\includegraphics[width=0.45\linewidth]{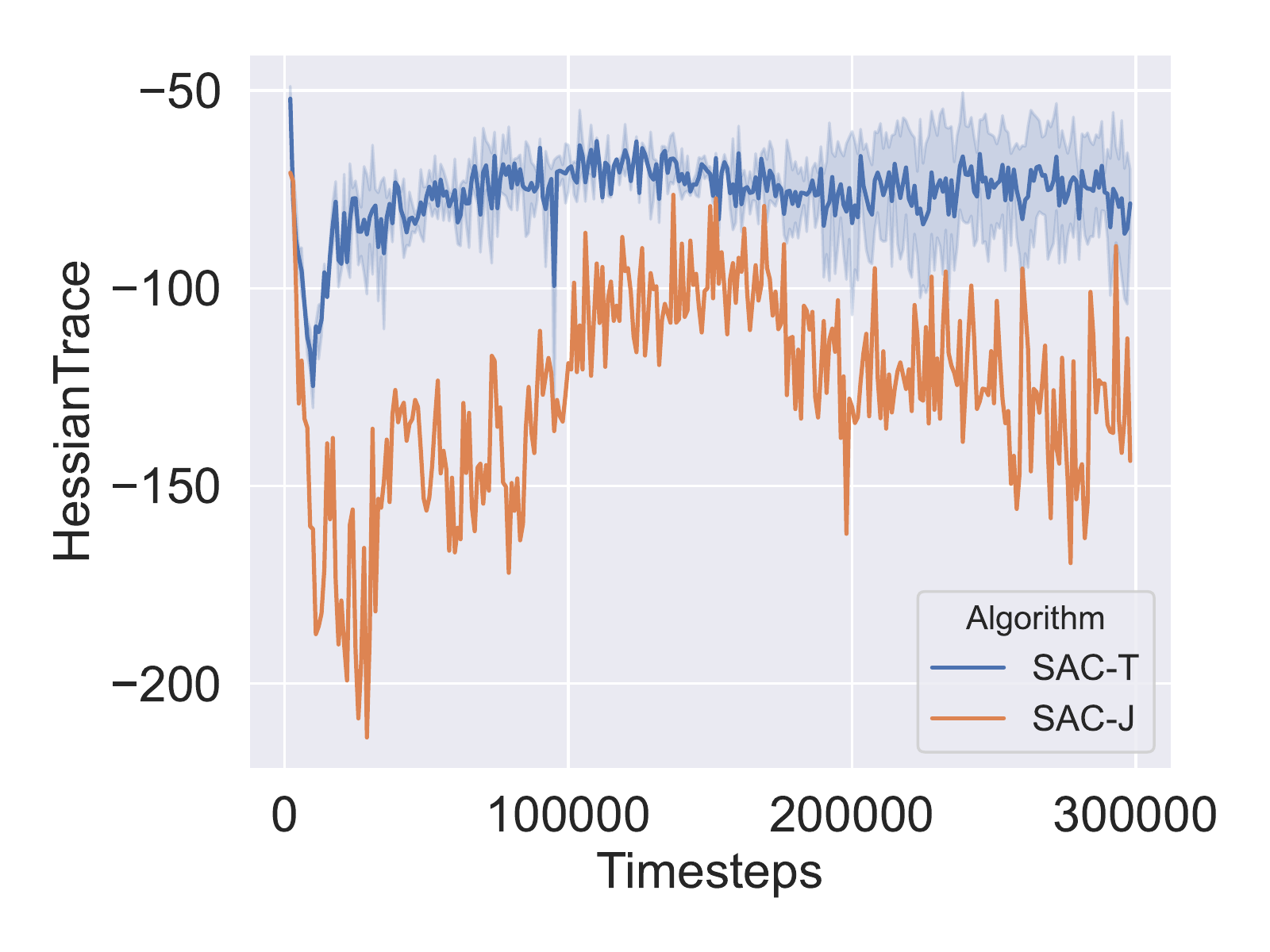}}\\
        \subfloat[Walker2D-v3 (Mujoco)]{\label{fig-walker}\includegraphics[width=0.45\linewidth]{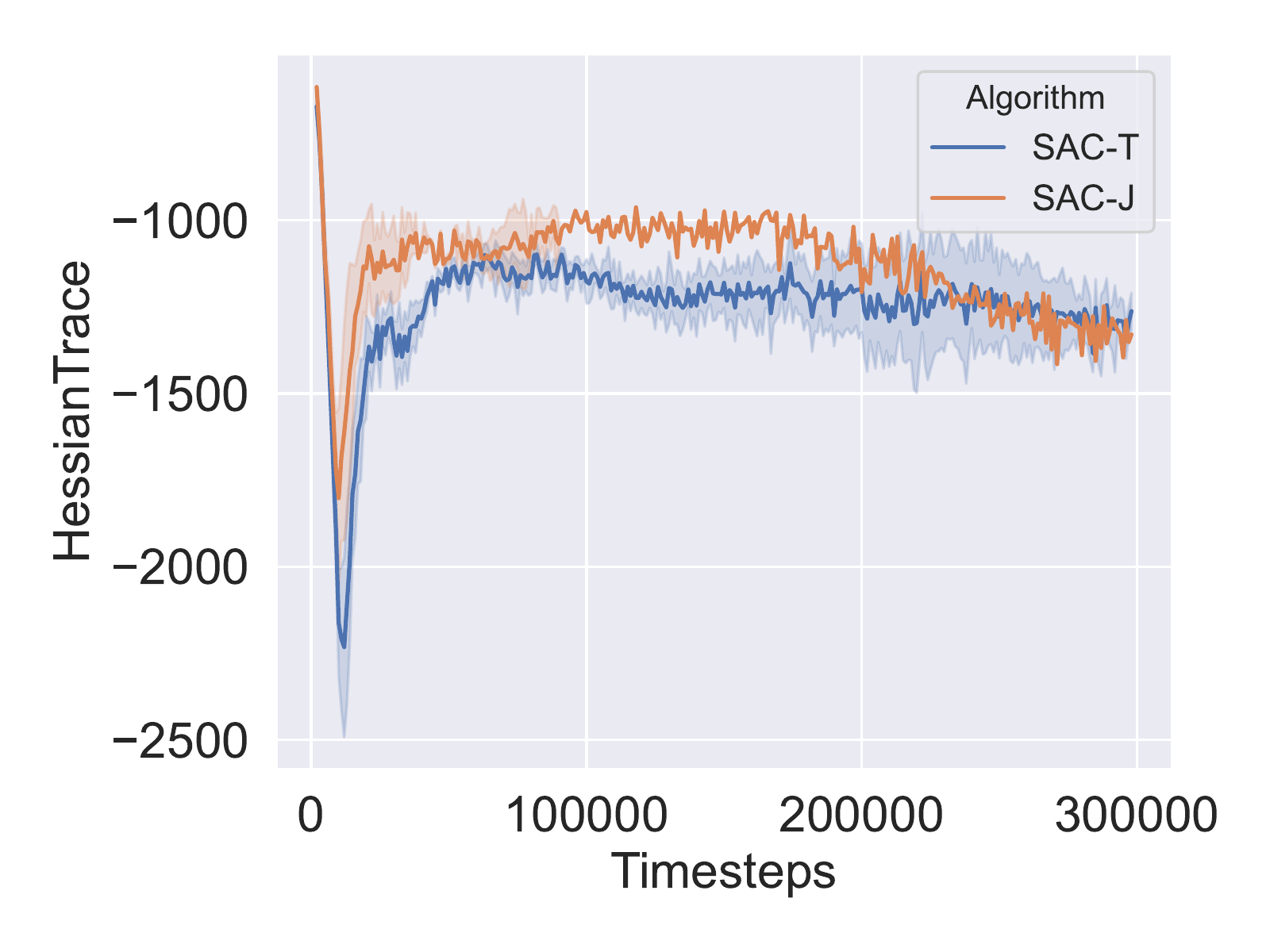}}
        \subfloat[Hopper-v0 (PyBullet)]{\label{fig-hopperPB}\includegraphics[width=0.45\linewidth]{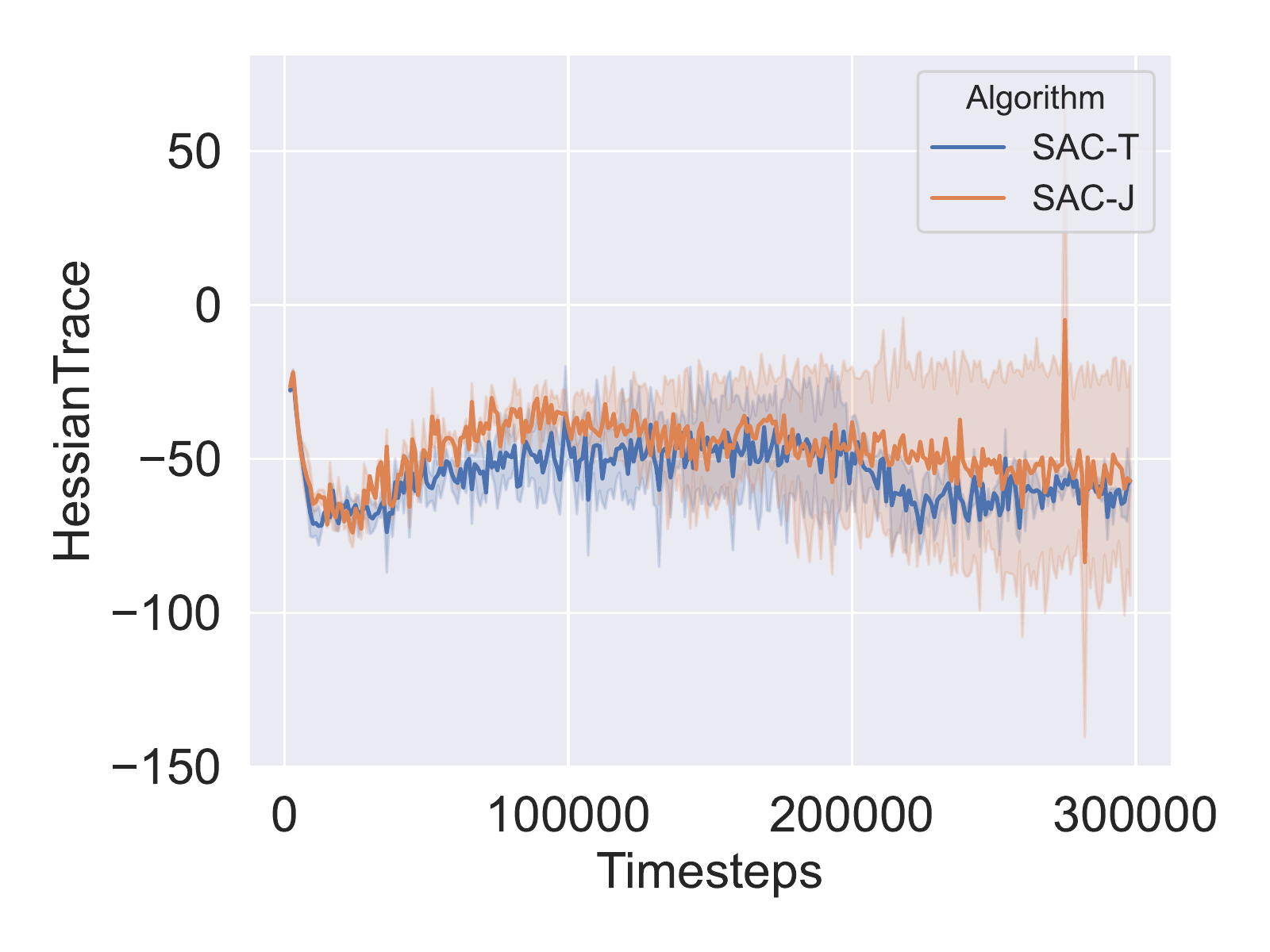}}
    \end{center}
    \caption{Hessian trace trend during the training process for SAC-J and SAC-T.}
    \label{fig:sac_hessian_trace_perf}
\end{figure*}

\begin{figure*}[!hbt]
    \begin{center}
        \subfloat[LunarLanderContinuous-v2]{\label{fig-ll}\includegraphics[width=0.45\linewidth]{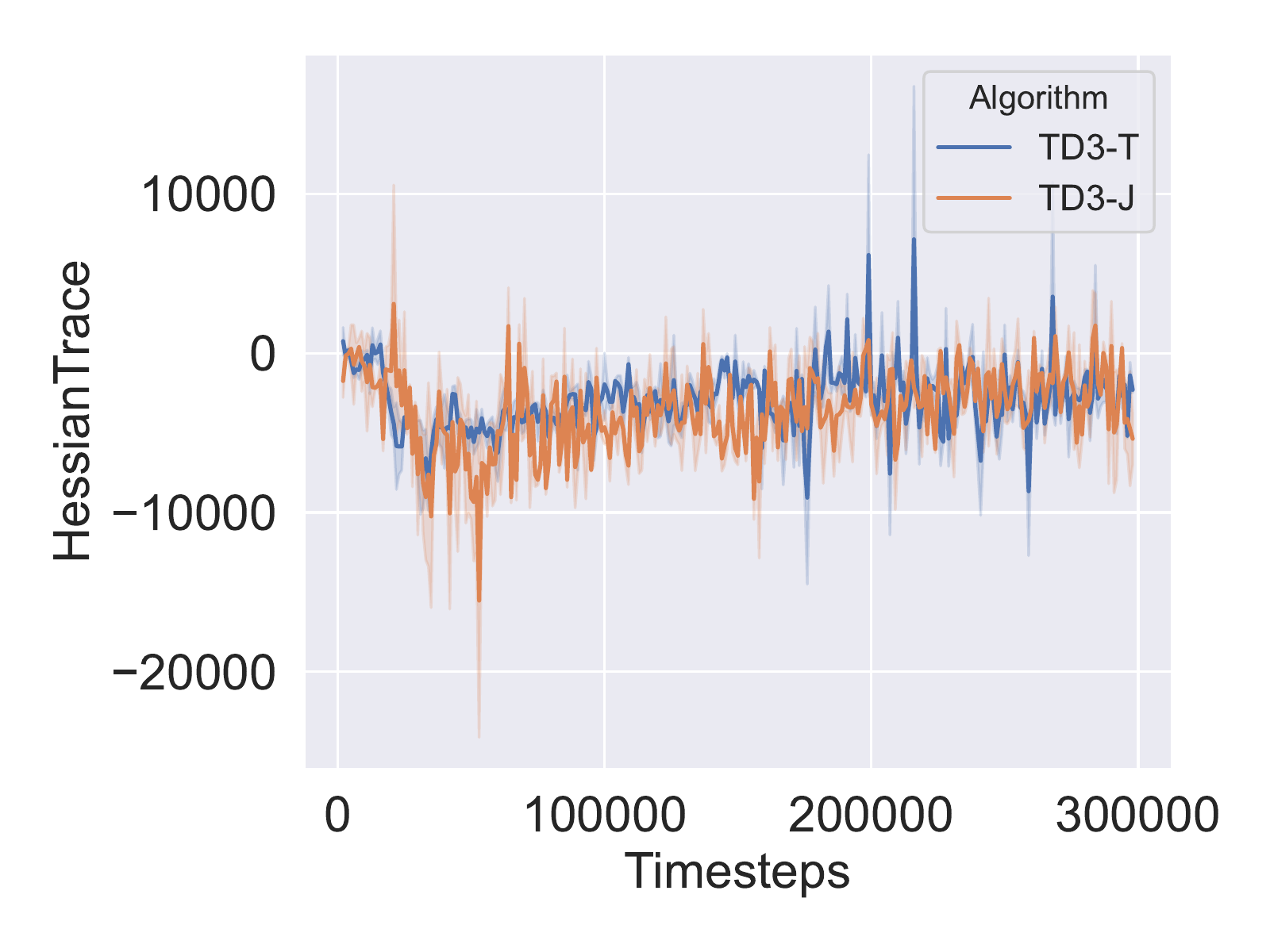}}
        \subfloat[Walker2D-v0 (PyBullet)]{\label{fig-walker}\includegraphics[width=0.45\linewidth]{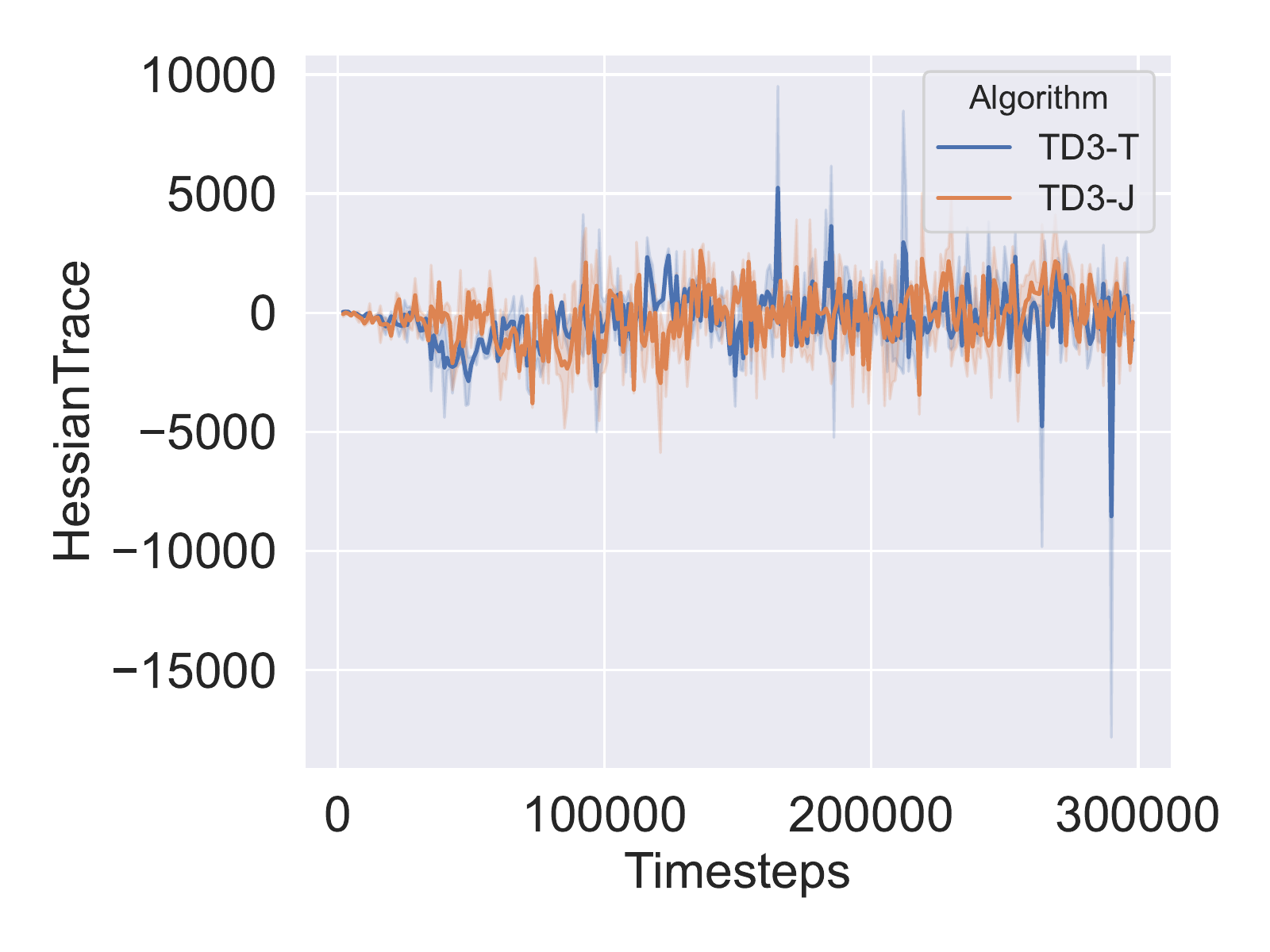}}
    \end{center}
    \caption{Hessian trace trend during the training process for TD3-J and TD3-T.}
    \label{fig:td3_hessian_trace_perf}
\end{figure*}

In this subsection, we experimentally show the effectiveness of using the proposed metric tensor DNN to learn $g_{ab}|_{\theta}$ with respect to any policy parameter $\theta$ so that $|Div(J^a)|_{\theta}|$ can be made closer to zero. For this purpose, we introduce a new quantity named the \emph{divergence ratio}, which is defined as the absolute ratio between the divergence of $J^a$ (i.e. $Div(J^a)|_{\theta}$ in the manifold $(\mathbb{R}^n,g_{ab})$) and the Hessian trace of the policy gradient. Note that the Hessian trace is the divergence of the policy gradient vector field in the Euclidean policy parametric space (i.e. the metric tensor field of the manifold is the identity metric tensor $\delta_{ab}$).

The divergence ratio quantifies the relative divergence changes upon extending the Euclidean policy parametric space into a generalized Riemannian manifold with the introduction of the metric tensor field $g_{ab}$. Specifically, whenever the divergence ratio is less than 1 and close to 0, the absolute divergence $Div(J^a)|_{\theta}$ in the manifold $(\mathbb{R}^n,g_{ab})$ is smaller than the absolute divergence in the Euclidean policy parametric space, implying that the policy gradient vector field becomes smoother in the manifold $(\mathbb{R}^n,g_{ab})$. As demonstrated by the experiment results reported in Section \ref{sub-sec-perf-comp}, this allows policy network training to be performed effectively and stably.

On the other hand, if the divergence ratio is above 1, it indicates that the policy gradient vector field becomes less smooth in the manifold $(\mathbb{R}^n,g_{ab})$. In this case, our metric tensor regularized policy gradient algorithms will resort to using normal policy gradients in the Euclidean policy parametric space to train the policy networks.


Figure~\ref{fig:sac_divergencesac_training_perf} presents the divergence ratios obtained by SAC-T during the training process on four benchmark problems. Evidenced by the figure, using the trained metric tensor DNN and the corresponding $g_{ab}$, SAC-T successfully reduces a significant portion of the divergence ratios to below 1 during the training process. As reported in Table~\ref{tab:divergence-ratio}, over 60\% of the divergence ratios obtained by SAC-T during policy training are less than 1 on all benchmark problems. This results demonstrates the effectiveness of our metric tensor regularization algorithm in training the proposed metric tensor DNN to achieve zero-divergence in the manifold $(\mathbb{R}^n,g_{ab})$.

In addition to the above analysis, we further present the Hessian trace obtained by SAC and TD3 on several benchmark problems respectively in Figures \ref{fig:sac_hessian_trace_perf} and \ref{fig:td3_hessian_trace_perf}. Interestingly, the two figures show that the Hessian trace obtained by using the same algorithm such as SAC-T can vary greatly on different benchmark problems. Meanwhile, even on the same benchmark problem, the Hessian traces produced by different algorithms such as SAC-T and TD3-T can be significantly different. Driven by this understanding, we believe the impact of Hessian trace on the performance of policy gradient algorithms should never be neglected. Our metric tensor regularized policy gradients present the first successful attempt in the literature towards utilizing and controlling the Hessian trace for effective and reliable training of policy networks.


\section{Conclusions}

In this paper, we studied policy gradient techniques for deep reinforcement learning. Motivated by the understanding that most of the existing policy gradient algorithms relied on the first-order policy gradient information to train policy networks, we aim to develop new mathematical tools and deep learning methods to effectively utilize and control Hessian information associated with the policy gradient in order to boost the performance of these algorithms. We focused on studying the Hessian trace as the key Hessian information, which gives the divergence of the policy gradient vector field in the Euclidean policy parametric space. In order to reduce the absolute divergence to zero so as to smoothen the policy gradient vector field, we successfully developed new mathematical tools, deep learning techniques and metric DNN architectures in this paper. Armed with these new technical developments, we have further proposed a new metric tensor regularized policy gradient algorithm based on SAC and TD3. The newly developed algorithm was further evaluated experimentally on several benchmark RL problems. Our experiment results confirmed that the new metric tensor regularized algorithm can significantly outperform its counterpart that does not use our regulization mechanism. Additional experiment results also confirmed that the trained metric tensor DNN in our algorithm can effectively reduce the absolute divergence towards zero in the Riemmanian manifold.







\bibliographystyle{named}
\bibliography{reference}

\newpage

\section*{Appendix A}

This appendix presents a proof of Proposition \ref{prop-1}. The divergence of $\mathrm{C}^{\infty}$ vector field $J^a|_{\theta}$ at any $\theta\in\mathbb{R}^n$ satisfies the equation below:
$$
Div(J^a)|_{\theta}=\nabla_a J^a = \frac{1}{\sqrt{|g|}} \sum_{\mu=1}^n \frac{\partial}{\partial \theta^{(\mu)}}\left( \sqrt{|g|} \vec{J}^{(\mu)} \right)
$$
where $g=det(G_{\theta})$. Following the specific structure of $G_{\theta}$ in \eqref{equ-G} and using the matrix determinant lemma \cite{press2007numerical},
$$
g=1+\vec{u}^T(\theta,\phi)\cdot \vec{u}(\theta,\phi) > 0
$$
Hence, $\sqrt{|g|}=\sqrt{g}$. Let
\begin{equation*}
\begin{split}
\epsilon&=\sum_{\mu=1}^n \frac{\partial}{\partial \theta^{(\mu)}}\left( \sqrt{|g|} \vec{J}^{(\mu)} \right) \\
&=\sqrt{g}\sum_{\mu=1}^n \left( \frac{\partial \vec{J}^{(\mu)}}{\partial \theta^{(\mu)}}+\frac{\vec{J}^{(\mu)}}{\sqrt{g}}\frac{\partial \sqrt{g}}{\partial \theta^{(\mu)}} \right)
\end{split}
\end{equation*}
Using Jacobi's formula \cite{bellman1997introduction} below
$$
\frac{\partial}{\partial \theta^{(\mu)}} det(G_{\theta}) = det(G_{\theta}) Tr\left(G_{\theta}^{-1} \frac{\partial G_(\theta)}{\partial \theta^{(\mu)}}\right),
$$
$\epsilon$ can be re-written as
$$
\epsilon = \sqrt{g} \sum_{\mu=1}^n \left( \frac{\partial \vec{J}^{(\mu)}}{\partial \theta^{(\mu)}} + \frac{\vec{J}^{(\mu)}}{2} Tr\left(G_{\theta}^{-1} \frac{\partial G_(\theta)}{\partial \theta^{(\mu)}}\right) \right)
$$
Notice that
$$
\frac{\partial G_{\theta}}{\partial \theta^{(\mu)}}=\left( \frac{\partial\vec{u}(\theta,\phi)}{\partial \theta^{(\mu)}} \right)\cdot \vec{u}(\theta,\phi)^T + \vec{u}(\theta,\phi) \cdot \left( \frac{\partial \vec{u}(\theta,\phi)}{\partial \theta^{(\mu)}} \right)^T
$$
Clearly there are two parts in the above equation. We refer to them respectively as $P1(\frac{\partial G_{\theta}}{\partial \theta^{(\mu)}})$ and $P2(\frac{\partial G_{\theta}}{\partial \theta^{(\mu)}})$. Using these notations,
\begin{equation*}
\begin{split}
G_{\theta}^{-1}P2\left(\frac{\partial G_{\theta}}{\partial \theta^{(\mu)}}\right) &= \vec{u}(\theta,\phi) \cdot \left( \frac{\partial \vec{u}(\theta,\phi)}{\partial \theta^{(\mu)}} \right)^T - \frac{ \vec{u}(\theta,\phi)\cdot\vec{u}(\theta,\phi)^T }{ 1+\vec{u}(\theta,\phi)^T\cdot\vec{u}(\theta,\phi) } \vec{u}(\theta,\phi) \left( \frac{\partial \vec{u}(\theta,\phi)}{\partial \theta^{(\mu)}} \right)^T \\
&=\frac{1}{\vec{u}(\theta,\phi)^T\cdot\vec{u}(\theta,\phi)} \vec{u}(\theta,\phi) \left( \frac{\partial \vec{u}(\theta,\phi)}{\partial \theta^{(\mu)}} \right)^T
\end{split}
\end{equation*}
Meanwhile,
\begin{equation*}
\begin{split}
G_{\theta}^{-1}P1\left(\frac{\partial G_{\theta}}{\partial \theta^{(\mu)}}\right) &=\left( \frac{\partial\vec{u}(\theta,\phi)}{\partial \theta^{(\mu)}} \right)\cdot \vec{u}(\theta,\phi)^T - \frac{\vec{u}(\theta,\phi)\cdot\vec{u}(\theta,\phi)^T}{1+\vec{u}(\theta,\phi)^T\cdot\vec{u}(\theta,\phi)} \left( \frac{\partial\vec{u}(\theta,\phi)}{\partial \theta^{(\mu)}} \right)\cdot \vec{u}(\theta,\phi)^T \\
&= \left( \frac{\partial\vec{u}(\theta,\phi)}{\partial \theta^{(\mu)}} \right)\cdot \vec{u}(\theta,\phi)^T - \frac{ \vec{u}(\theta,\phi)^T\cdot \frac{\partial \vec{u}(\theta,\phi)}{\partial \theta^{(\mu)}} }{1+\vec{u}(\theta,\phi)^T\cdot\vec{u}(\theta,\phi)} \vec{u}(\theta,\phi)\cdot \vec{u}(\theta,\phi)^T
\end{split}
\end{equation*}
Subsequently,
\begin{equation*}
\begin{split}
Tr\left( G_{\theta}^{-1}\frac{\partial G_{\theta}}{\partial \theta^{(\mu)}} \right)&=Tr\left( \frac{1}{\vec{u}(\theta,\phi)^T\cdot\vec{u}(\theta,\phi)} \vec{u}(\theta,\phi) \left( \frac{\partial \vec{u}(\theta,\phi)}{\partial \theta^{(\mu)}} \right)^T \right) + Tr\left( \left( \frac{\partial\vec{u}(\theta,\phi)}{\partial \theta^{(\mu)}} \right)\cdot \vec{u}(\theta,\phi)^T  \right) \\
&\ \ \ - Tr\left( \frac{ \vec{u}(\theta,\phi)^T\cdot \frac{\partial \vec{u}(\theta,\phi)}{\partial \theta^{(\mu)}} }{1+\vec{u}(\theta,\phi)^T\cdot\vec{u}(\theta,\phi)} \vec{u}(\theta,\phi)\cdot \vec{u}(\theta,\phi)^T \right) \\
&= \frac{ \vec{u}(\theta,\phi)^T \cdot \left( \frac{\partial \vec{u}(\theta,\phi)}{\partial \theta^{(\mu)}} \right) }{\vec{u}(\theta,\phi)^T\cdot\vec{u}(\theta,\phi)} + \vec{u}(\theta,\phi)^T\cdot \left( \frac{\partial\vec{u}(\theta,\phi)}{\partial \theta^{(\mu)}} \right) \\
&\ \ \ - \frac{ \vec{u}(\theta,\phi)^T \cdot \vec{u}(\theta,\phi) }{1+\vec{u}(\theta,\phi)^T\cdot\vec{u}(\theta,\phi)} \vec{u}(\theta,\phi)^T\cdot \frac{\partial \vec{u}(\theta,\phi)}{\partial \theta^{(\mu)}} \\
&= \frac{2}{1+\vec{u}(\theta,\phi)^T\cdot\vec{u}(\theta,\phi)}\vec{u}(\theta,\phi)^T\cdot \frac{\partial \vec{u}(\theta,\phi)}{\partial \theta^{(\mu)}}
\end{split}
\end{equation*}
Using the above equation, we have
$$
\epsilon=\sqrt{g} \sum_{\mu=1}^n \left( \frac{\partial \vec{J}^{(\mu)}}{\partial \theta^{(\mu)}} + \frac{\vec{J}^{(\mu)}}{1+\vec{u}(\theta,\phi)^T\cdot \vec{u}(\theta,\phi)}\sum_{\nu=1}^n \vec{u}^{(\nu)}(\theta)\frac{\partial \vec{u}^{(\nu)}(\theta)}{\partial \theta^{(\mu)}} \right)
$$
This proves the claim in Proposition \ref{prop-1} below
\begin{equation*}
\begin{split}
Div(J^a)|_{\theta}&=\frac{1}{\sqrt{g}}\epsilon\\
&=\sum_{\mu=1}^n \left( \frac{\partial \vec{J}^{(\mu)}}{\partial \theta^{(\mu)}} + \frac{\vec{J}^{(\mu)}}{1+\vec{u}(\theta,\phi)^T\cdot \vec{u}(\theta,\phi)}\sum_{\nu=1}^n \vec{u}^{(\nu)}(\theta)\frac{\partial \vec{u}^{(\nu)}(\theta)}{\partial \theta^{(\mu)}} \right)
\end{split}    
\end{equation*}

\section*{Appendix B}

This appendix presents a proof of Proposition \ref{prop-2}. For any $A\in\mathcal{SO}(n)$,
$$
\exp(A)=I_n + A + \frac{1}{2!}A^2+\frac{1}{3!}A^3+\ldots
$$
We can conduct SVD decomposition of A such that
$$
A=U\cdot\Sigma\cdot V^T
$$
with $U$ and $V$ being $n\times n$ unitary matrices. $\Sigma=Diag(\vec{sigma})$ is a diagonal matrix. Therefore,
$$
\exp(A)=I_n+U\cdot\Sigma\cdot V^T + \frac{1}{2!}(U\cdot\Sigma\cdot V^T)(U\cdot\Sigma\cdot V^T)+\frac{1}{3!}(U\cdot\Sigma\cdot V^T)(U\cdot\Sigma\cdot V^T)(U\cdot\Sigma\cdot V^T)+\ldots
$$
Note that $A^T=-A$, hence
$$
(U\cdot\Sigma\cdot V^T)^T=-U\cdot\Sigma\cdot V^T=V\cdot\Sigma\cdot U^T
$$
Consequently, $\forall k\geq 1$
\begin{equation*}
A^k=\left\{
\begin{array}{ll}
(-1)^{k/2} U\cdot\Sigma^k\cdot U^T, & k \text{ is even;} \\
(-1)^{(K+1)/2} V\cdot\Sigma^k\cdot U^T, & k \text{ is odd.}
\end{array}
\right.
\end{equation*}
In line with the above, we have
\begin{equation*}
\begin{split}
\exp(A) &= \sum_{k=0}^{\infty} \frac{(-1)^k}{(2k)!} U\cdot\Sigma^{2k}\cdot U^T - \sum_{k=0}^{\infty} \frac{(-1)^k}{(2k+1)!}V\cdot\Sigma^{2k+1}\cdot U^T \\
&= U\cdot \left[
\begin{array}{ccc}
cos(\vec{\sigma}^{(1)}) & 0 & 0 \\
0 & \ddots & 0 \\
0 & 0 & cos(\vec{\sigma}^{(n)})
\end{array}
\right]\cdot U^T -  V\cdot \left[
\begin{array}{ccc}
sin(\vec{\sigma}^{(1)}) & 0 & 0 \\
0 & \ddots & 0 \\
0 & 0 & sin(\vec{\sigma}^{(n)})
\end{array}
\right]\cdot U^T \\
&= U\cdot \Sigma_c \cdot U^T - V\cdot \Sigma_s \cdot U^T
\end{split}
\end{equation*}
This proves Proposition \ref{prop-2}.

\section*{Appendix C}

This appendix presents a proof of Proposition \ref{prop-3}. Following the assumption that $\exp(A)=\hat{\Omega}\cdot\Sigma_c\cdot\hat{\Omega}^T-\hat{\Phi}\cdot\Sigma_s\cdot\hat{\Omega}^T$, for any vector $\vec{a}$, we have
$$
\exp(A)\cdot\vec{a}=\hat{\Omega}\cdot\Sigma_c\cdot\hat{\Omega}^T\cdot\vec{a}-\hat{\Phi}\cdot\Sigma_s\cdot\hat{\Omega}^T\cdot\vec{a}
$$
Using Fourier transformation, we can re-write vector $\vec{a}$ in the Fourier series form below:
$$
\vec{a}^{(j)}=\eta_0 + \sqrt{\frac{2}{n}} \sum_{i=1}^n \left[
\eta_i cos\left( \frac{2\pi i}{n} j \right) + \tilde{\eta}_i sin\left( \frac{2\pi i}{n} j \right)
\right]
$$
Hence,
$$
\hat{\Omega}^T\cdot\vec{a} = \left[
\begin{array}{c}
\eta_1 \\
\vdots \\
\eta_n
\end{array}
\right]
$$
where $\eta_i=(\vec{\hat{\Omega}}^{(i)})^T\cdot \vec{a}$. Subsequently,
$$
\Sigma_c\cdot\hat{\Omega}^T\cdot\vec{a} = \left[
\begin{array}{c}
cos(\vec{\sigma}^{(1)})\eta_1 \\
\vdots \\
cos(\vec{\sigma}^{(n)})\eta_n
\end{array}
\right] \text{ and }
\Sigma_s\cdot\hat{\Omega}^T\cdot\vec{a} = \left[
\begin{array}{c}
sin(\vec{\sigma}^{(1)})\eta_1 \\
\vdots \\
sin(\vec{\sigma}^{(n)})\eta_n
\end{array}
\right]
$$
Therefore,
\begin{equation*}
\begin{split}
\hat{\Omega}\cdot\Sigma_c\cdot\hat{\Omega}^T\cdot\vec{a} &= [\vec{\hat{\Omega}}^{(1)},\ldots,\vec{\hat{\Omega}}^{(1)}]\cdot \left[
\begin{array}{c}
cos(\vec{\sigma}^{(1)})\eta_1 \\
\vdots \\
cos(\vec{\sigma}^{(n)})\eta_n
\end{array}
\right] \\
&= \sum_{j=1}^n cos(\vec{\sigma}^{(j)}) \eta_j \vec{\hat{\Omega}}^{(j)}
\end{split}
\end{equation*}

\begin{equation*}
\begin{split}
\hat{\Phi}\cdot\Sigma_c\cdot\hat{\Omega}^T\cdot\vec{a} &= [\vec{\hat{\Phi}}^{(1)},\ldots,\vec{\hat{\Phi}}^{(1)}]\cdot \left[
\begin{array}{c}
sin(\vec{\sigma}^{(1)})\eta_1 \\
\vdots \\
sin(\vec{\sigma}^{(n)})\eta_n
\end{array}
\right] \\
&= \sum_{j=1}^n sin(\vec{\sigma}^{(j)}) \eta_j \vec{\hat{\Phi}}^{(j)}
\end{split}
\end{equation*}
We can now re-write $\exp(A)\cdot\vec{a}$ as
\begin{equation*}
\begin{split}
\exp(A)\cdot\vec{a}&=\sum_{i=1}^n \eta_i \left( cos(\vec{\sigma}^{(i)}) \vec{\hat{\Omega}}^{(i)} - sin(\vec{\sigma}^{(i)}) \vec{\hat{\Phi}}^{(i)} \right) \\
&= \sqrt{\frac{2}{n}} \sum_{i=1}^n \eta_i \left[
\begin{array}{l}
cos(\vec{\sigma}^{(i)}) cos\left( \frac{2\pi i}{n} j \right)|_{j=0} - sin(\vec{\sigma}^{(i)}) sin\left( \frac{2\pi i}{n} j \right)|_{j=n-1} \\
\vdots \\
cos(\vec{\sigma}^{(i)}) cos\left( \frac{2\pi i}{n} j \right)|_{j=0} - sin(\vec{\sigma}^{(i)}) sin\left( \frac{2\pi i}{n} j \right)|_{j=n-1}
\end{array}
\right] \\
&= \sqrt{\frac{2}{n}} \sum_{i=1}^n \eta_i \left[
\begin{array}{c}
cos\left( \frac{2\pi i}{n} j + \vec{\sigma}^{(i)} \right)|_{j=0} \\
\vdots \\
cos\left( \frac{2\pi i}{n} j + \vec{\sigma}^{(i)} \right)|_{j=n-1}
\end{array}
\right]
\end{split}    
\end{equation*}
In other words,
$$
(\exp{A}\cdot\vec{a})^{(j)}=\sqrt{\frac{2}{n}} \sum_{i=1}^n 
\eta_i cos\left( \frac{2\pi i}{n} j + \vec{\sigma}^{(i)} \right)
$$
This concludes that, when applying $exp(A)$ to vector $\vec{a}$, it will lead to independent phase shifts of the frequency components of $\vec{a}$. In other words, rotating the $i$-th frequency component is equivalent to a phase shift of $\vec{\sigma}^{(i)}$ for that frequency component. This ends the proof of Proposition \ref{prop-3}.

\section*{Appendix D}

This appendix presents a proof of Proposition \ref{prop-4}. Any geodesic that passes through $\theta$ in manifold $(\mathbb{R}^n,g_{ab})$ and has $J^a|_{\theta}$ as its tangent vector at $\theta$ can be uniquely determined by the \emph{geodesic equation} below \cite{kreyszig2013differential}:
$$
\frac{\mathrm{d}^2 \theta^{(\mu)}(t)}{\mathrm{d} t^2}+\sum_{\nu=1}^n \sum_{\delta=1}^n \Gamma^{\mu}_{\nu,\delta} \frac{\mathrm{d} \theta^{(\nu)}(t)}{\mathrm{d} t} \frac{\mathrm{d} \theta^{(\delta)}(t)}{\mathrm{d} t}=0, \mu=1,\ldots,n
$$
where $t$ stands for the geodesic parameter such that $\theta^{(\mu)}(0)=\theta^{(\mu)}$. $\Gamma^{\mu}_{\nu,\delta}$ or $\Gamma^a_{b,c}$ in the abstract index notation is the \emph{Christoff symbol}. Therefore,
$$
\frac{\mathrm{d}^2 \theta^{(\mu)}(t)}{\mathrm{d} t^2} = -\sum_{\nu=1}^n \sum_{\delta=1}^n \Gamma^{\mu}_{\nu,\delta} \frac{\mathrm{d} \theta^{(\nu)}(t)}{\mathrm{d} t} \frac{\mathrm{d} \theta^{(\delta)}(t)}{\mathrm{d} t}
$$
subject to the conditions
$$
\left( \frac{\mathrm{d} \theta^{(\nu)}(t)}{\mathrm{d}t}\right)|_{t=0}=\vec{J}^{(\nu)}, \nu=1,\ldots,n
$$
Hence, updating $\theta$ along the direction of the geodesic can be approximated by the following learning rule:
$$
\theta^{(\mu)}\leftarrow \theta^{(\mu)} + \alpha \left(\frac{\mathrm{d} \theta^{(\mu)}(t)}{\mathrm{d} t}\right)|_{t=0}  - \alpha \Delta t \sum_{\nu=1}^n \sum_{\delta=1}^n \left[ \Gamma^{\mu}_{\nu,\delta} \left(\frac{\mathrm{d} \theta^{(\nu)}(t)}{\mathrm{d} t}\right)|_{t=0} \left( \frac{\mathrm{d} \theta^{(\delta)}(t)}{\mathrm{d} t} \right)|_{t=0} \right]
$$
where $\alpha$ is the learning rate. $\Delta t$ refers to a small increment of the geodesic parameter at $t=0$. In view of the above, the geodesic regularized policy gradient can be approximated as
$$
\vec{T}^{(\mu)}\approx \vec{J}^{(\mu)}- \Delta t \sum_{\nu=1}^n \sum_{\delta=1}^n \Gamma^{\mu}_{\nu,\delta} \vec{J}^{(\nu)} \vec{J}^{(\delta)}
$$
Because
\begin{equation*}
\begin{split}
\Gamma^{\delta}_{\mu,\nu}&=\Gamma^c_{a,b} (\mathrm{d}\theta^{\delta})_c \left( \frac{\partial}{\partial \theta^{(\mu)}} \right)^a \left( \frac{\partial}{\partial \theta^{(\nu)}} \right)^b \\
&=\frac{1}{2} \sum_{\rho=1}^n g^{\delta,\rho}\left(
\frac{\partial g_{\nu,\rho}}{\partial \theta^{(\mu)}} + \frac{\partial g_{\mu,\rho}}{\partial \theta^{(\nu)}} \right) - \frac{1}{2} \sum_{\rho=1}^n g^{\delta,\rho} \left(\frac{\partial g_{\mu,\nu}}{\partial \theta^{(\rho)}}
\right)
\end{split}
\end{equation*}
We can study the two summations in the above equation separately. Let us denote
$$
P1(\Gamma^{\delta}_{\mu,\nu})=\frac{1}{2} \sum_{\rho=1}^n g^{\delta,\rho}\left(
\frac{\partial g_{\nu,\rho}}{\partial \theta^{(\mu)}} + \frac{\partial g_{\mu,\rho}}{\partial \theta^{(\nu)}} \right)
$$
$$
P2(\Gamma^{\delta}_{\mu,\nu})=\frac{1}{2} \sum_{\rho=1}^n g^{\delta,\rho} \left(\frac{\partial g_{\mu,\nu}}{\partial \theta^{(\rho)}}
\right)
$$
Consequently,
\begin{equation*}
\begin{split}
\sum_{\mu=1}^n \sum_{\nu=1}^n P1(\Gamma^{\delta}_{\mu,\nu}) \frac{\mathrm{d} \theta^{(\mu)}(t)}{\mathrm{d} t} \frac{\mathrm{d} \theta^{(\nu)}(t)}{\mathrm{d} t} &= \frac{1}{2} \sum_{\mu=1}^n \sum_{\nu=1}^n \sum_{\rho=1}^n g^{\delta,\rho} \left(
\frac{\partial g_{\nu,\rho}}{\partial \theta^{(\mu)}} + \frac{\partial g_{\mu,\rho}}{\partial \theta^{(\nu)}} \right) \frac{\mathrm{d} \theta^{(\mu)}(t)}{\mathrm{d} t} \frac{\mathrm{d} \theta^{(\nu)}(t)}{\mathrm{d} t} \\
&= \frac{1}{2} \sum_{\rho=1}^n g^{\delta,\rho} \left(
\sum_{\mu=1}^n \sum_{\nu=1}^n \left(
\frac{\partial g_{\nu,\rho}}{\partial \theta^{(\mu)}} + \frac{\partial g_{\mu,\rho}}{\partial \theta^{(\nu)}} \right) \frac{\mathrm{d} \theta^{(\mu)}(t)}{\mathrm{d} t} \frac{\mathrm{d} \theta^{(\nu)}(t)}{\mathrm{d} t}
\right)
\end{split}
\end{equation*}
Note that
$$
\sum_{\mu=1}^n \sum_{\nu=1}^n \left(
\frac{\partial g_{\nu,\rho}}{\partial \theta^{(\mu)}} + \frac{\partial g_{\mu,\rho}}{\partial \theta^{(\nu)}} \right) \frac{\mathrm{d} \theta^{(\mu)}(t)}{\mathrm{d} t} \frac{\mathrm{d} \theta^{(\nu)}(t)}{\mathrm{d} t}=2  \sum_{\nu=1}^n
\left( \sum_{\mu=1}^n \frac{\partial g_{\nu,\rho}}{\partial \theta^{(\mu)}} \frac{\mathrm{d} \theta^{(\mu)}(t)}{\mathrm{d} t} \right) \frac{\mathrm{d} \theta^{(\nu)}(t)}{\mathrm{d} t}
$$
In particular, $\sum_{\mu=1}^n \frac{\partial g_{\nu\rho}}{\partial \theta^{(\mu)}} \frac{\mathrm{d} \theta^{\mu}}{\mathrm{d} t}$ captures the change of $g_{ab}$ along the direction of the geodesic. In view of this, since $g_{ab}$ is expected to change smoothly and stably along the geodesic, i.e.
$$
\Delta t \sum_{\mu=1}^n \frac{\partial g_{\nu,\rho}}{\partial \theta^{(\mu)}} \frac{\mathrm{d} \theta^{(\mu)}(t)}{\mathrm{d} t}\approx -\zeta_1 \cdot g_{\nu,\rho}, \zeta_1>0
$$
Using the above,
\begin{equation*}
\begin{split}
\Delta t\sum_{\mu=1}^n \sum_{\nu=1}^n P1(\Gamma^{\delta}_{\mu,\nu}) \left(\frac{\mathrm{d} \theta^{\mu}(t)}{\mathrm{d} t}\right)|_{t=0} \left(\frac{\mathrm{d} \theta^{\nu}(t)}{\mathrm{d} t}\right)|_{t=0} & \approx \sum_{\rho=1}^n g^{\delta,\rho} \sum_{\nu=1}^n (-\zeta_1 \cdot g_{\rho,\nu}) \left(\frac{\mathrm{d} \theta^{\nu}(t)}{\mathrm{d} t}\right)|_{t=0}\\
& = -\zeta_1 \left(\frac{\mathrm{d} \theta^{\delta}(t)}{\mathrm{d} t}\right)|_{t=0} \\
& = -\zeta_1 \vec{J}^{(\delta)}
\end{split}
\end{equation*}
Accordingly,
$$
\vec{T}^{(\sigma)}\approx \vec{J}^{(\sigma)} + \zeta_1 \vec{J}^{(\sigma)} + \frac{\Delta t}{2} \sum_{\rho=1}^n g^{\delta,\rho} \sum_{\mu=1}^n \sum_{\nu=1}^n \left(\frac{\partial g_{\mu,\nu}}{\partial \theta^{(\rho)}}
\right) \left(\frac{\mathrm{d} \theta^{\mu}(t)}{\mathrm{d} t}\right)|_{t=0} \left(\frac{\mathrm{d} \theta^{\nu}(t)}{\mathrm{d} t}\right)|_{t=0}
$$
Let $\zeta_2=\frac{\Delta t}{2}$, we have
$$
\vec{T}^{(\sigma)}\approx \vec{J}^{(\sigma)}+\zeta_1 \vec{J}^{(\sigma)} + \zeta_2 \sum_{\rho=1}^n g^{\delta,\rho} \sum_{\mu=1}^n \sum_{\nu=1}^n \left(\frac{\partial g_{\mu,\nu}}{\partial \theta^{(\rho)}}
\right) \vec{J}^{(\mu)} \vec{J}^{(\nu)}
$$
This proves Proposition \ref{prop-4}. We can also re-write the above equation in the form of a matrix expression below for easy implementation by a deep learning library.
$$
\vec{T}\approx (1+\zeta_1) \vec{J} + \zeta_2 G_{\theta}^{-1} \cdot \nabla_{\theta} \left( NoGrad(\vec{J})^T \cdot G_{\theta}\cdot NoGrad(\vec{J})\right)
$$
Here, $NoGrad(\vec{J})$ indicates that vector $\vec{J}$ will not participate in the gradient calculation. $\nabla_{\theta}$ stands for the normal gradient operator with respect to $\theta$. Using the approximated $\vec{T}$, we can build a new learning rule below:
$$
\theta \leftarrow \theta + \alpha \vec{T}
$$
In line with this learning rule, $\frac{\zeta_2}{1+\zeta_1}>0$ will be treated as a hyper-parameter of the $g_{ab}$ regularization algorithm.

\end{document}